\documentclass[10pt,twocolumn,letterpaper]{article}

\usepackage{iccv}
\usepackage{times}
\usepackage{epsfig}
\usepackage{graphicx}
\usepackage{amsmath}
\usepackage{amssymb}

\usepackage{multirow}
\usepackage{booktabs}
\usepackage{arydshln} 
\usepackage{subcaption}

\usepackage[breaklinks=true,bookmarks=false]{hyperref}

\iccvfinalcopy 

\def\iccvPaperID{8067} 

\begin{document}

\title{Can Targeted Adversarial Examples Transfer When the Source and Target Models Have No Label Space Overlap?}

\author{Nathan Inkawhich$^{1}$, Kevin J Liang$^{2}$, Jingyang Zhang$^{1}$, Huanrui Yang$^{1}$, Hai Li$^{1}$ and Yiran Chen$^{1}$\\
$^{1}$Duke University \qquad $^{2}$Facebook AI\\
{\tt\small nathan.inkawhich@duke.edu}
}

\maketitle
\ificcvfinal\thispagestyle{empty}\fi

\begin{abstract}
We design blackbox transfer-based targeted adversarial attacks for an environment where the attacker's source model and the target blackbox model may have disjoint label spaces and training datasets. This scenario significantly differs from the ``standard'' blackbox setting, and warrants a unique approach to the attacking process. Our methodology begins with the construction of a class correspondence matrix between the whitebox and blackbox label sets. During the online phase of the attack, we then leverage representations of highly related proxy classes from the whitebox distribution to fool the blackbox model into predicting the desired target class. Our attacks are evaluated in three complex and challenging test environments where the source and target models have varying degrees of conceptual overlap amongst their unique categories. Ultimately, we find that it is indeed possible to construct targeted transfer-based adversarial attacks between models that have non-overlapping label spaces! We also analyze the sensitivity of attack success to properties of the clean data. Finally, we show that our transfer attacks serve as powerful adversarial priors when integrated with query-based methods, markedly boosting query efficiency and adversarial success.
\end{abstract}

\newcommand{\fdanlxent}[1]{\textit{FDA}$^{(#1)}$\textit{+xent}}
\newcommand{\fdanl}[1]{\textit{FDA}$^{(#1)}$}

\vspace{-5mm}
\section{Introduction} \label{sec:intro}
\vspace{-1mm}

\begin{figure}[t]
\centering
    \includegraphics[width=0.9\linewidth,trim={0 0 0 0},clip]{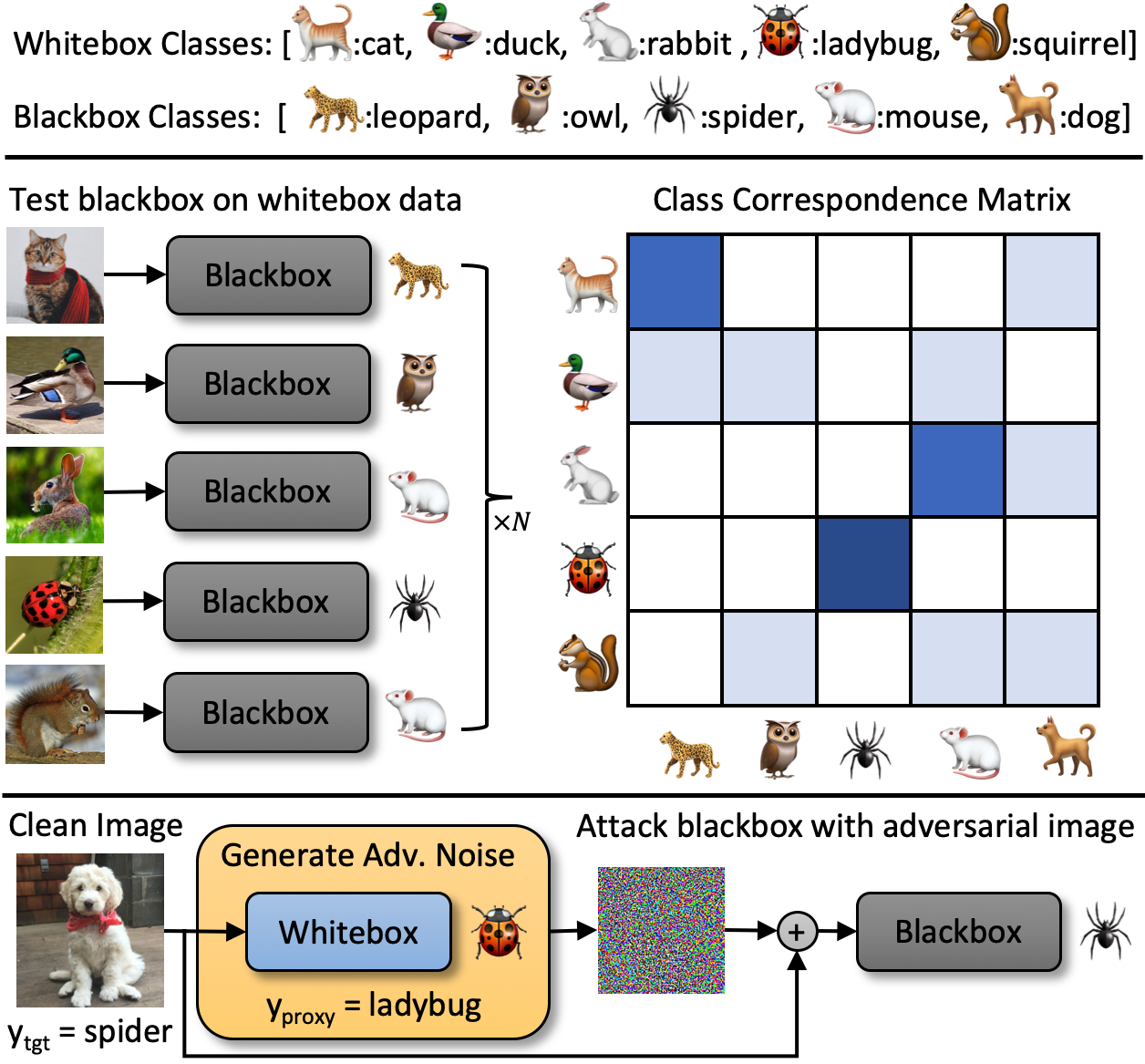}
\vspace{-2mm}
\caption{Overview of attack setup and procedure.}
\label{fig:overview}
\vspace{-6mm}
\end{figure}

The intrigue of blackbox adversarial attacks is that only limited knowledge of the target model is assumed, making them more difficult but also more practical. Of the several blackbox techniques, transfer-based attacks have become popular because they do not rely on repetitively querying the target model during attack generation \cite{mim_attack, sgm_iclr20, inkawhich_cvpr19, inkawhich_neurips20, Naseer0KKP19}. 
While modern query-based methods are powerful \cite{ChengDPSZ19, IlyasEAL18, TuTC0ZYHC19}, they are susceptible to simple input rate throttling defenses because they often require thousands of inquires per adversarial example. 
Despite this no-query advantage, a glaring weakness in the design and evaluation of current transfer attacks is the implicit assumption that \textit{the attacker's source model is trained on the exact same dataset as the target blackbox model!} This oversight is largely due to the common practice of testing transfers exclusively between pre-trained models from the same benchmark dataset (e.g., MNIST, CIFAR-10, and ImageNet) and has likely lead to overly optimistic estimates of transferability that may not translate to real-world settings, in which an attacker may not have access to the target model's data.

In this work, we explore the potency of transfer attacks in a much more restrictive setting, where this shared dataset assumption is eliminated. 
We draw inspiration from the recent findings of Inkawhich~\textit{et al.}~\cite{inkawhich_neurips20}, who observe that targeted transfer attacks maintain potency when there is zero training data overlap, yet significant label space overlap between the source and target models.
Taking it a step further, here we investigate the feasibility of targeted transfer attacks when not only is there zero training data overlap, but also \textit{no label space overlap} between the source and target model training distributions.
Upon initial consideration, the premise of this disjoint label space transfer scenario may seem absurd. The formulation of most contemporary targeted attacks (including \cite{inkawhich_neurips20}) boils down to maximizing the likelihood of a target class. However, if the attacker's whitebox model does not have the target class in its label space, how can targeted transfer be optimized for? Our primary goal is thus to probe whether or not it is possible to achieve attack success in such settings, and if so, to what extent.

Figure~\ref{fig:overview} shows a basic overview of our attack setup and procedure. Given a whitebox and blackbox model that are trained on unique class sets, we propose first constructing a correspondence matrix from the blackbox's predictions on samples from the whitebox distribution, to identify any relationships that exist between classes across label spaces. This is followed by an adversarial noise generation phase, which leverages highly-related proxy classes from the whitebox's label space to induce the desired targeted response by the blackbox.
Intuitively, our proposed attack methodology seeks to exploit a shared feature set, along with a set of potentially shared adversarial sub-spaces~\cite{LeeICLR21, Naseer0KKP19} between the two training distributions, whose presence is suggested by high responses in the correspondence matrix. 

As a part of our core experiments, we design several unique test environments of varying difficulty that re-define the whitebox and blackbox training distributions. Ultimately, we find that in each environment a variety of target classes in the blackbox model's label space can be reliably attacked. We also observe the particular adversarial noise-generation algorithm can have a large impact on transferability and error rates. Specifically, attacks that leverage the intermediate feature space of the whitebox \cite{inkawhich_iclr20, inkawhich_neurips20} appear to be more potent than attacks that use the output layer \cite{mim_attack, ensemble_attack1}. Finally, we perform experiments where small amounts of queries to the blackbox model are allowed during attack generation, and find that our transfer attacks can be used as powerful priors to reduce the number of queries necessary to achieve targeted attack success. Overall, we demonstrate that it is in fact possible to achieve targeted transfer between models with no label space overlap and outline a novel method for exploiting such a vulnerability.

\vspace{-1mm}
\section{Related work}
\vspace{-1mm}
\textbf{``Standard'' transfer attacks.} Within the topic area of blackbox transfer-based adversarial attacks \cite{KurakinNeurIPSAdv}, there are several predominant methodologies for creating adversarial noise, and one overwhelmingly popular way to evaluate attack effectiveness. Most works closely follow the design of whitebox attacks \cite{GoodfellowSS14, MadryMSTV18, KurakinGB17}, and incorporate additional optimization tricks and regularization terms in an effort to reduce over-fitting to the whitebox model's decision boundaries and architecture \cite{mim_attack, ti_attack, sinim_attack, sgm_iclr20, di_attack, ensemble_attack1, abs-2012-11207}. Uniquely, feature space attacks create adversarial noise to manipulate the intermediate layer representations of the whitebox, and in many cases achieve state-of-the-art transferability~\cite{RozsaGB17, inkawhich_cvpr19, HuangKGHBL19, inkawhich_iclr20, lu2020enhancing, LiGC20_eccv, inkawhich_neurips20}. The ``standard'' approach to evaluate these attacks is to transfer between different model architectures trained on the same benchmark datasets (e.g., CIFAR-10 or ImageNet). Thus, the measurement of transferability refers to attack potency across deep neural network (DNN) architectures, but carries the (significant) implicit assumption that the models have been trained on the exact same data distribution and label set.

\textbf{Reducing attacker knowledge.}
Some recent works have considered blackbox transfer attacks in more strict settings, where the shared dataset assumption is relaxed. Both Bose~\textit{et al.}~\cite{BoseGBCVLH20} and Li, Guo and Chen~\cite{LiGC20} develop un-targeted attacks in the non-interactive blackbox (NoBox) threat model, which specifies that the attacker: (1) may not issue queries to the target model; and (2) has a ``reference dataset'' which is sampled from the same distribution as the target model's training set (so, the label space turns out to be the same). By contrast, we focus on targeted attacks and do not assume access to \textit{any} data sampled from the blackbox distribution when training our whitebox models. Naseer~\textit{et al.}~\cite{Naseer0KKP19} train a generator network to produce image-agnostic un-targeted attacks between models with no label space overlap; however, do not consider targeted attacks within this threat model. Lastly, Inkawhich~\textit{et al.}~\cite{inkawhich_neurips20} construct ``cross-distribution'' transfer scenarios with zero training data overlap but significant (yet incomplete) label space overlap between the whitebox and blackbox models. Through experiments, they confirm that targeted adversarial examples can be constructed for the overlapping classes, but do not attempt to attack any non-overlapping classes.

\vspace{-1mm}
\section{Methodology} \label{sec:method}
\vspace{-1mm}
\subsection{Establishing class correspondences}
\vspace{-1mm}
The first phase of our attack methodology is to model the relationships between the classes of the whitebox and blackbox. Note, this step is unique to our disjoint label space setting, as in previous works the class relationships are explicitly encoded by the intersection of the label sets. We accomplish this task by forward passing a limited amount of data from each category of the whitebox model's training distribution through the target model and recording the predictions in a class correspondence matrix. Intuitively, this matrix describes how the blackbox model responds to the presence of whitebox data features, which will become relevant when constructing adversarial perturbations. 

Although the whitebox data is technically out-of-distribution (OOD) w.r.t.~the blackbox's training set, the closed-world assumption commonly made during DNN training means the blackbox has poor confidence calibration on OOD data \cite{BendaleB16}, allowing for this type of analysis. As an attacker, we make note of any hot-spots in the correspondence matrix, which represent promising (target, proxy) relationships to potentially exploit (e.g., (\texttt{leopard}, \texttt{cat}) and (\texttt{spider}, \texttt{ladybug}) in Figure~\ref{fig:overview}). Importantly, we consider this step to occur ``offline'' w.r.t.~the actual attack, as it is only necessary to do once (before computing any adversarial examples) and does not involve inputting adversarially perturbed data through the target model.

It is worth noting a few basic assumptions made in this step. First, we assume the adversary has at least some samples from the training distribution of their own whitebox models, as opposed to only having a pre-trained model. We believe this to be a realistic assumption, as in most cases the attacker may either be using a model trained on a widely available benchmark dataset, or would have had to train the model themselves. 
The second assumption is that the attacker is allowed to issue a small amount of queries to the target model and receive predictions (only the predicted class is necessary, not the full probability vectors). Since this is the most basic function of an ``oracle'' blackbox model, we believe this to be reasonable.

\vspace{-1mm}
\subsection{Computing adversarial perturbations}
\vspace{-1mm}
The second phase of our methodology is to construct the targeted adversarial perturbations. We start with a clean image $x$ and target class $y_{tgt}$ from the blackbox distribution.
The goal is then to compute adversarial noise $\delta$ using the whitebox model such that the blackbox model classifies the adversarial example $x+\delta$ as $y_{tgt}$ (as shown in the bottom of Figure~\ref{fig:overview}). To do this, we first index the class correspondence matrix with $y_{tgt}$ to find a highly correlated proxy class $y_{proxy}$ from the whitebox distribution that can be used to represent $y_{tgt}$ during attack generation. Intuitively, since $y_{tgt}$ and $y_{proxy}$ images are interpreted similarly by the blackbox model, imparting the features of $y_{proxy}$ onto $x$ through an adversarial noising process may in-turn cause the blackbox model to misclassify the adversarial example as $y_{tgt}$. We explore two fundamentally different approaches for optimizing the adversarial noise: decision-space methods and feature-space methods. 

\textbf{Decision-space Attack.}
Decision-space methods work to directly manipulate the output-layer signals of the whitebox model. Often, this is accomplished by minimizing the classification loss w.r.t.~a designated target class (in our case $y_{proxy}$). Thus, the adversarial perturbation is quasi-optimal for the whitebox, as it is designed to traverse exact decision boundaries. Let $f$ represent the whitebox model and $f(x)$ be the predicted probability distribution over its set of classes. We use the powerful Targeted Momentum Iterative Method (\textit{TMIM}) \cite{mim_attack} as a representative of decision-space attacks, whose optimization objective is 
\begin{equation}
    \min_{\delta \in \mathcal{S}(x;\epsilon)} H \big (f(x+\delta), y_{proxy} \big ). 
    \label{eqn:tmim}
\end{equation}
Here, $H(f(x+\delta), y_{proxy})$ is the cross-entropy between the whitebox's predicted distribution and $y_{proxy}$. The constraint $\mathcal{S}(x;\epsilon)$ defines an allowable perturbation set w.r.t.~$x$, often to keep $\delta$ imperceptible.
Optimizing this objective results in adversarial examples capable of fooling the whitebox model into predicting $y_{proxy}$ with high confidence.
We hypothesize that by pushing $x+\delta$ into a high probability region of $y_{proxy}$ in the whitebox, these decision-based attacks may in-turn cause the blackbox to regard $x+\delta$ as $y_{tgt}$.

\textbf{Feature-space Attack.}
As the name suggests, feature space methods compute adversarial noise using the intermediate feature information of the whitebox \cite{inkawhich_iclr20, inkawhich_neurips20}. Rather than optimizing to explicitly cross decision boundaries, these methods make the adversarial examples ``look like'' the target class (or in our case $y_{proxy}$) in feature space. In this work, we use the multi-layer Feature Distribution Attack (\textit{FDA}) \cite{inkawhich_neurips20} as our representative feature space method. For setup, we first train the necessary auxiliary feature distribution models for each proxy class at a specified set of whitebox model layers $\mathcal{L}=\{\ell_1,\ldots,\ell_N\}$. Note, the class $c$, layer $\ell$ auxiliary model inputs the whitebox's layer $\ell$ feature map $f_\ell(x)$, and outputs the probability that it is from an input of class $c$ (i.e., it outputs $p(y=c|f_{\ell}(x))$).

Using the trained auxiliary models, the \textit{FDA} attack objective function can then be defined as
\begin{equation}
    \max_{\delta \in \mathcal{S}(x;\epsilon)} L_{FDA}(f, x, y_{proxy}, \delta, \mathcal{L}, \eta),
\end{equation}
where 
\begin{equation}
    \begin{split}
        L&_{FDA}(f, x, y, \delta, \mathcal{L}, \eta) =\\ 
        & \frac{1}{|\mathcal{L}|} \sum_{\ell \in \mathcal{L}} p(y | f_\ell(x+\delta)) + \eta \frac{\left\Vert f_\ell(x+\delta) - f_\ell(x) \right\Vert _2}{\left\Vert f_\ell(x) \right\Vert _2}.
    \end{split}
\end{equation}
\noindent
By maximizing $L_{FDA}$, the adversarial noise: (1) maximizes the likelihood that intermediate features from across the whitebox's feature hierarchy belong to the proxy class; and (2) enforces that the perturbed image's feature map is significantly different from the clean image's feature map, as measured by a normalized $L_2$ distance. 

Our intuition for why this method has promise stems from the discussed potential overlap of features in the whitebox and blackbox data distributions.
Particularly, if there exists strong (target, proxy) relationships in the class correspondence matrix, we posit that this is clear evidence that the blackbox model has learned features that appear in the whitebox data distribution. Thus, the \textit{FDA} attack is well-phrased to manipulate any shared features that may exist between a given (target, proxy) class pair. Further, we hypothesize that since the perturbation objective of \textit{FDA} is detached from the exact decision boundary structure of the whitebox (which is irrelevant to the blackbox model anyway), it may yield higher transferability because the optimization is focused on the feature compositions.

\vspace{-1mm}
\section{Disjoint label space transfer environments} \label{sec:envs}
\vspace{-1mm}
Since we are not executing transfers in standard benchmark settings, we carefully design novel test environments to evaluate the efficacy and scalability of our attacks across a spectrum of realistic settings. We consider three distinct ``cross-distribution'' transfer scenarios. The first two, named Disjoint ImageNet Subsets (DINS) Test 1 and Test 2, are manually constructed by partitioning ImageNet \cite{DengDSLL009} into challenging subsets. The third scenario involves transfers between ImageNet and Places365 \cite{zhou2017places} models. 

\textbf{DINS Test 1 and 2.}
To assemble the DINS Test 1 and 2 environments, we first establish 30 distinct super-classes that are comprised of five individual ImageNet classes each. For example, we create a \texttt{bird} class by aggregating the ImageNet classes [~\textit{10:brambling; 11:goldfinch; 12:house-finch; 13:junco; 14:indigo-bunting}~], and a \texttt{big-cat} class by aggregating [~\textit{286:cougar; 287:lynx; 289:snow-leopard; 290:jaguar; 292:tiger}~]. The full list is shown in Appendix A. The super-classes are then partitioned as shown in Table~\ref{tab:class_splits} to create the two DINS environments. Each is comprised of non-overlapping 15-class ``A'' and ``B'' splits, to be referred to as Test 1.A/B and Test 2.A/B. In our experiments, the blackbox target models in each test are trained on the ``B'' splits, while the attacker's whitebox models are trained on the ``A'' splits. To be clear, we only intend to attack between ``A'' and ``B'' models under the same environment, i.e., Test 1.A$\rightarrow$Test 1.B and Test 2.A$\rightarrow$Test 2.B.

\begin{table}[t]
\centering
\caption{Disjoint ImageNet Subsets Test 1 and Test 2}
\vspace{-2mm}
\resizebox{0.99\linewidth}{!}{
\begin{tabular}{cccc}
\toprule
\multicolumn{2}{c}{DINS Test 1}    & \multicolumn{2}{c}{DINS Test 2}    \\ \cmidrule(lr){1-2} \cmidrule(lr){3-4}
A             & B             & A             & B             \\ \cmidrule(lr){1-2} \cmidrule(lr){3-4}
\texttt{fish}          & \texttt{crab}          & \texttt{dog-hound}     & \texttt{house-cat}     \\
\texttt{bird}          & \texttt{butterfly}     & \texttt{dog-terrier}   & \texttt{big-cat}       \\
\texttt{lizard}        & \texttt{snake}         & \texttt{dog-spaniel}   & \texttt{lizard}        \\
\texttt{spider}        & \texttt{beetle}        & \texttt{dog-retriever} & \texttt{snake}         \\
\texttt{dog-hound}     & \texttt{dog-terrier}   & \texttt{insect}        & \texttt{fish}          \\
\texttt{dog-spaniel}   & \texttt{dog-retriever} & \texttt{beetle}        & \texttt{bird}          \\
\texttt{house-cat}     & \texttt{big-cat}       & \texttt{butterfly}     & \texttt{spider}        \\
\texttt{insect}        & \texttt{fungus}        & \texttt{train}         & \texttt{small-vehicle} \\
\texttt{boat}          & \texttt{train}         & \texttt{instrument}    & \texttt{large-vehicle} \\
\texttt{small-vehicle} & \texttt{large-vehicle} & \texttt{boat}          & \texttt{computer}      \\
\texttt{mustelids}     & \texttt{big-game}      & \texttt{turtle}        & \texttt{big-game}      \\
\texttt{turtle}        & \texttt{monkey}        & \texttt{crab}          & \texttt{mustelids}    \\
\texttt{drinkware}     & \texttt{clothing}      & \texttt{drinkware}     & \texttt{sports-ball}   \\
\texttt{fruit}         & \texttt{sports-ball}   & \texttt{fruit}         & \texttt{clothing}      \\
\texttt{instrument}    & \texttt{computer}      & \texttt{monkey}        & \texttt{fungus}        \\ \bottomrule
\end{tabular}
}
\label{tab:class_splits}
\vspace{-4mm}
\end{table}

We remark that the DINS tests are created to represent different difficulty levels. DINS Test 1 illustrates an intuitively more promising transfer case, because there are some obvious conceptual overlaps between the classes in Test 1.A and 1.B that may lead to natural (target, proxy) relationships. For example, both contain the general concepts of dogs, cats and vehicles. However, the challenge is that both have different supports (i.e., underlying ImageNet classes) for what makes up a dog, cat and vehicle. DINS Test 2 represents an intuitively harder transfer environment, where there is much less conceptual overlap between the label spaces of the whitebox (Test 2.A) and blackbox (Test 2.B). Notice, all four of the dog sub-breeds are in Test 2.A, while all of the cat and vehicle categories are in Test 2.B. Here, it is much less obvious which Test 2.B classes can be targeted with the available proxy classes in Test 2.A. 

\textbf{ImageNet to Places365.}
Our third transfer scenario is principally created to evaluate the scalability of our attacks to more complex environments. We consider a situation where the attacker has whitebox access to ImageNet models and wishes to create targeted attacks for Places365 blackbox models. The additional complexity in this experiment comes from two primary sources. First, the sheer increase in the number of classes in both the whitebox and blackbox label spaces: ImageNet has 1000 classes and Places365 has 365 classes. Second, there is generally a finer granularity between the categories in both distributions. For example, instead of classifying at the stratum of dogs, cats, fish, etc. (as we do in the DINS tests), these more complex models have to classify between highly nuanced object sub-categories (e.g., there are over 110 dog breeds in ImageNet). From an attacking perspective, such an increase in complexity and granularity may make it more difficult to reliably target individual classes in the Places365 label space. Finally, we note that the ImageNet and Places365 datasets technically have about 10 classes worth of label space overlap, which represents a very small percentage of the union of categories. However, we intend to transfer across both overlapping and non-overlapping sectors of the label space.

\begin{figure*}[t]
\centering
    \includegraphics[width=0.9\linewidth,trim={0 0 0 0},clip]{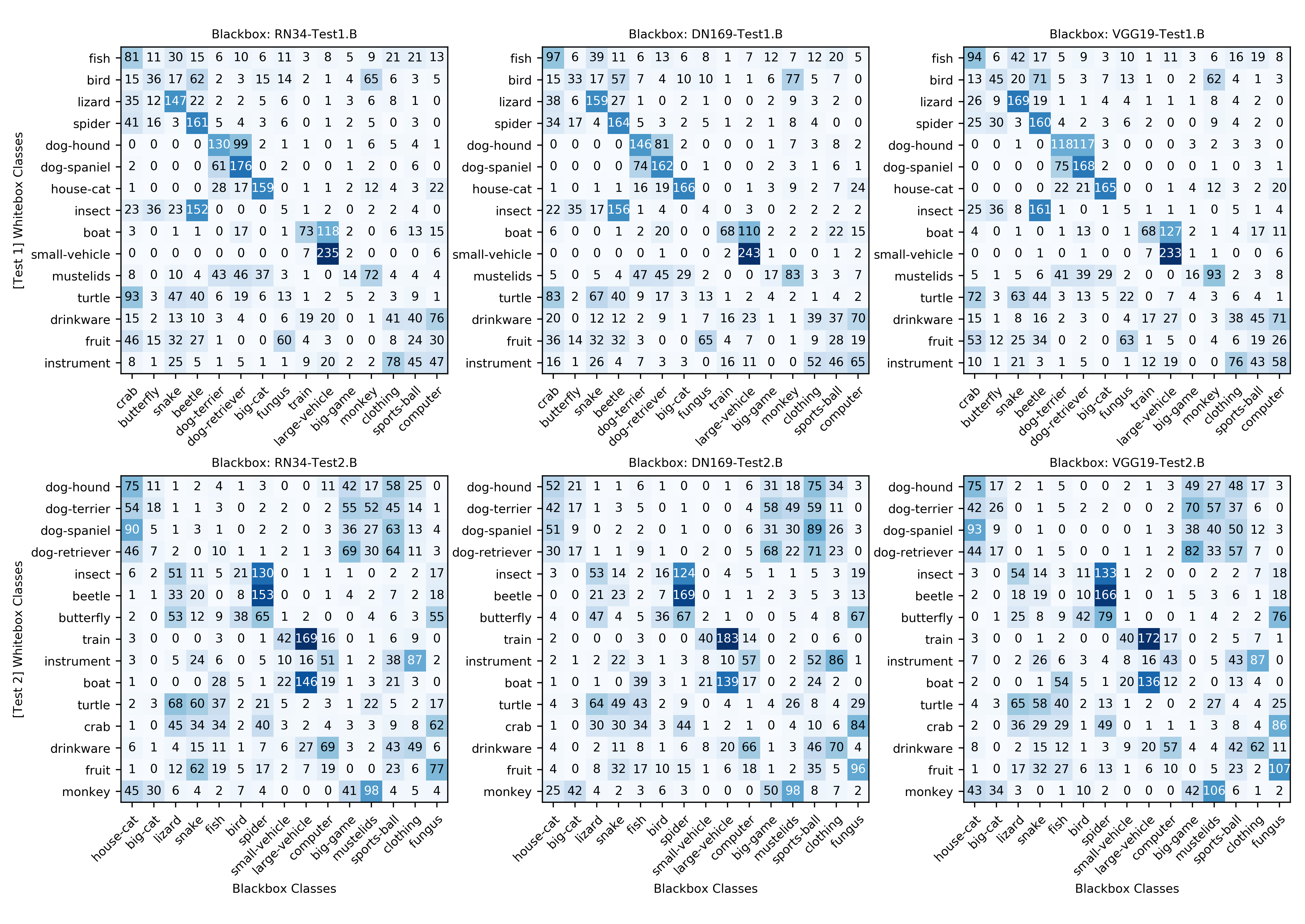}
    \vspace{-6mm}
\caption{Class correspondence matrices for the DINS Test 1 (top row) and DINS Test 2 (bottom row) environments.}
\vspace{-4mm}
\label{fig:class_correlations}
\end{figure*}

\vspace{-1mm}
\section{Experiments}
\vspace{-1mm}
Our experiments are split into a few main sections. First, we describe the setup and results of attacking in the DINS Test 1 and 2 environments. We then perform an analysis of source class impact, and show an extension of the transfer attacks to an environment where limited queries to the blackbox may be allowed. Finally, we discuss the results of transferring in the ImageNet to Places365 environment.

\vspace{-1mm}
\subsection{Experimental setup} \label{sec:setup}
\vspace{-1mm}
On the Test 1.B and Test 2.B splits, we train ResNet-34 (RN34)~\cite{HeZRS16}, ResNet-152 (RN152)~\cite{HeZRS16}, DenseNet-169 (DN169)~\cite{HuangLMW17}, VGG19bn \cite{SimonyanZ14a}, MobileNetv2 (MNv2)~\cite{SandlerHZZC18}, and ResNeXt-50 (RXT50)~\cite{XieGDTH17} models to be used as blackboxes. We then train ResNet-50 (RN50) and DenseNet-121 (DN121) models on both Test 1.A and Test 2.A to act as whiteboxes. A variety of blackbox architectures, including ones that are not from the same architectural family as the whitebox models, are purposely considered to produce a quality estimate of transferability.

When optimizing the adversarial objectives, both \textit{TMIM} and \textit{FDA} attacks use an $L_{\infty}~\epsilon=16/255$ noise constraint and iterate for 10 perturbing iterations while including momentum \cite{ti_attack, sgm_iclr20, inkawhich_neurips20}. We opt to use an ensemble variant of each attack method, where adversarial noise is optimized using both the RN50 and DN121 whiteboxes concurrently \cite{ensemble_attack1}. Note, the \textit{FDA} attack method involves a tuning step to select which layers are included in the attacking layer set \cite{inkawhich_neurips20}. Critically, we tune the attack layers only using models trained on the A splits (i.e., RN50-Test1.A$\leftrightarrow$DN121-Test1.A and RN50-Test2.A$\leftrightarrow$DN121-Test2.A). In this way, no extra queries to the blackbox models are required for attack hyper-parameter tuning.

Finally, to measure attack potency we consider two metrics: error-rate (error) and targeted-success-rate (tSuc). Because we enforce that all ``clean'' starting images are correctly classified by the target blackbox model, both error and tSuc have values in $[0, 1]$ (where numbers closer to one indicate higher success). Specifically, error measures the rate at which the blackbox model misclassifies the adversarial examples, and tSuc is the rate at which the blackbox model misclassifies the adversarial example as the specified target class. Even though our attacks are primarily optimized for tSuc, the rate of incidental errors can still be of interest to an attacker, especially because very limited options for un-targeted attacks exist in this threat model \cite{Naseer0KKP19}. See Appendix B for more details regarding experimental setup.

\vspace{-1mm}
\subsection{Measuring class correlations}
\vspace{-1mm}
\begin{table*}[t]
\centering
\caption{Transfers in the DINS Test 1 environment (notation = error / tSuc)}
\vspace{-2mm}
\resizebox{0.8\textwidth}{!}{
\begin{tabular}{cccccccccc}
\toprule
                               &                                &        & \multicolumn{6}{c}{Blackbox Models (Test 1.B)}       &                              \\ \cmidrule(lr){4-9}
Target (Test 1.B)               & Proxy (Test 1.A)                & Attack & RN34        & RN152       & DN169       & VGG19bn     & MNv2        & RXT50   & avg.    \\ \midrule
\multirow{2}{*}{\textbf{\texttt{large-vehicle}}} & \multirow{2}{*}{\textbf{\texttt{small-vehicle}}} & TMIM   & 25.3 / 4.3  & 21.1 / 3.5  & 25.6 / 7.1  & 26.0 / 6.9  & 30.5 / 7.3  & 29.5 / 8.1  & 26.3 / 6.2 \\
                               &                                & FDA    & 74.5 / 62.5 & 65.7 / 48.8 & 87.1 / 83.4 & 82.1 / 76.5 & 74.5 / 64.2 & 70.7 / 56.6 & \textbf{75.8 / 65.3} \\[0.1mm] \cdashline{1-10} \\[-3mm]
\multirow{2}{*}{\textbf{\texttt{snake}}}        & \multirow{2}{*}{\textbf{\texttt{lizard}}}        & TMIM   & 30.5 / 11.5 & 26.0 / 10.7 & 30.8 / 12.7 & 32.7 / 9.9  & 36.7 / 14.9 & 32.7 / 14.1 & 31.6 / 12.3 \\
                               &                                & FDA    & 63.7 / 52.2 & 64.9 / 56.0 & 79.0 / 70.6 & 77.3 / 47.7 & 68.8 / 55.8 & 65.1 / 56.5 & \textbf{69.8 / 56.5} \\[0.1mm] \cdashline{1-10} \\[-3mm]
\multirow{2}{*}{\textbf{\texttt{dog}-\textit{any}}}       & \multirow{2}{*}{\textbf{\texttt{dog-spaniel}}}   & TMIM   & 23.0 / 9.0  & 22.7 / 10.5 & 31.3 / 20.5 & 31.9 / 18.9 & 29.7 / 13.4 & 26.9 / 10.5 & 27.6 / 13.8 \\
                               &                                & FDA    & 71.8 / 64.2 & 59.8 / 51.9 & 85.0 / 80.6 & 89.2 / 84.7 & 79.9 / 75.3 & 66.9 / 58.3 & \textbf{75.4 / 69.2} \\[0.1mm] \cdashline{1-10} \\[-3mm]
\multirow{2}{*}{\textbf{\texttt{big-cat}}}       & \multirow{2}{*}{\textbf{\texttt{house-cat}}}     & TMIM   & 29.9 / 13.1 & 23.1 / 7.6  & 29.5 / 12.5 & 33.2 / 13.6 & 33.7 / 10.7 & 29.5 / 10.1 & 29.8 / 11.3 \\
                               &                                & FDA    & 74.9 / 59.4 & 63.1 / 46.0 & 82.4 / 66.3 & 83.2 / 55.4 & 75.2 / 51.9 & 65.6 / 45.4 & \textbf{74.1 / 54.1} \\[0.1mm] \cdashline{1-10} \\[-3mm]
\multirow{2}{*}{\textbf{\texttt{beetle}}}        & \multirow{2}{*}{\textbf{\texttt{insect}}}        & TMIM   & 28.5 / 10.2 & 26.9 / 9.2  & 30.2 / 9.5  & 30.0 / 6.9  & 31.3 / 4.7  & 30.5 / 8.3  & 29.6 / 8.1 \\
                               &                                & FDA    & 66.3 / 41.9 & 62.6 / 40.1 & 76.6 / 35.2 & 75.1 / 28.3 & 65.1 / 22.1 & 62.5 / 24.7 & \textbf{68.0 / 32.0} \\[0.1mm] \cdashline{1-10} \\[-3mm]
\multirow{2}{*}{\texttt{beetle}}        & \multirow{2}{*}{\texttt{spider}}        & TMIM   & 22.6 / 3.5  & 20.2 / 2.5  & 22.1 / 2.9  & 26.5 / 1.7  & 28.7 / 1.7  & 26.7 / 2.6  & 24.5 / 2.5 \\
                               &                                & FDA    & 44.6 / 9.8  & 36.6 / 3.9  & 51.4 / 7.2  & 55.4 / 3.5  & 46.9 / 3.5  & 40.7 / 6.4  & \textbf{45.9 / 5.7} \\[0.1mm] \cdashline{1-10} \\[-3mm]
\multirow{2}{*}{\texttt{large-vehicle}} & \multirow{2}{*}{\texttt{boat}}          & TMIM   & 25.2 / 0.6  & 19.8 / 0.5  & 23.1 / 0.7  & 26.0 / 1.6  & 29.1 / 1.7  & 27.9 / 1.2  & 25.2 / 1.1 \\
                               &                                & FDA    & 62.1 / 0.9  & 48.4 / 1.1  & 64.8 / 0.9  & 62.5 / 6.6  & 58.2 / 2.2  & 63.7 / 0.7  & \textbf{59.9 / 2.1} \\[0.1mm] \cdashline{1-10} \\[-3mm]
\multirow{2}{*}{\textbf{\texttt{crab}}}          & \multirow{2}{*}{\textbf{\texttt{fish}}}          & TMIM   & 24.3 / 4.9  & 22.8 / 6.0  & 24.8 / 7.5  & 27.2 / 4.9  & 28.0 / 6.2  & 28.1 / 5.4  & 25.9 / 5.8 \\
                               &                                & FDA    & 54.9 / 20.2 & 50.2 / 13.9 & 71.7 / 34.5 & 75.2 / 38.7 & 60.6 / 21.6 & 55.3 / 21.0 & \textbf{61.3 / 25.0} \\[0.1mm] \cdashline{1-10} \\[-3mm]
\multirow{2}{*}{\texttt{monkey}}        & \multirow{2}{*}{\texttt{mustelids}}    & TMIM   & 24.0 / 4.4  & 19.8 / 3.2  & 25.0 / 6.0  & 26.2 / 3.7  & 28.8 / 6.0  & 26.0 / 3.5  & 25.0 / 4.5 \\
                               &                                & FDA    & 60.1 / 17.7 & 52.4 / 10.1 & 69.5 / 11.0 & 73.4 / 21.4 & 68.3 / 13.8 & 57.0 / 7.9  & \textbf{63.4 / 13.7} \\[0.1mm] \cdashline{1-10} \\[-3mm]
\multirow{2}{*}{\texttt{clothing}}      & \multirow{2}{*}{\texttt{instrument}}    & TMIM   & 25.4 / 9.1  & 19.0 / 5.1  & 23.9 / 7.0  & 26.1 / 10.6 & 28.6 / 9.3  & 25.8 / 6.7  & 24.8 / 8.0 \\
                               &                                & FDA    & 40.8 / 12.5 & 33.3 / 9.3  & 44.4 / 6.7  & 46.0 / 8.9  & 47.4 / 11.1 & 43.3 / 5.1  & \textbf{42.5 / 8.9} \\[0.1mm] \cdashline{1-10} \\[-3mm]
\multirow{2}{*}{\textbf{\texttt{crab}}}          & \multirow{2}{*}{\textbf{\texttt{turtle}}}        & TMIM   & 27.9 / 5.3  & 23.4 / 6.8  & 28.0 / 8.0  & 29.8 / 7.2  & 31.3 / 6.7  & 29.1 / 5.4  & 28.2 / 6.6 \\
                               &                                & FDA    & 61.3 / 10.8 & 49.4 / 17.6 & 70.3 / 36.3 & 76.0 / 49.1 & 58.8 / 19.6 & 58.2 / 17.2 & \textbf{62.3 / 25.1} \\ \bottomrule
\end{tabular}
}
\label{tab:test1_results}
\vspace{-3mm}
\end{table*}

Figure~\ref{fig:class_correlations} shows the class correspondence matrices for the DINS Test 1 and 2 environments. The top row of subplots is the result of forward passing Test 1.A data through the RN34, DN169, and VGG19bn Test1.B models. The bottom row is the result of forward passing Test 2.A data through the RN34, DN169 and VGG19bn Test2.B models. From these matrices, our first observation is that in both environments there exist many strong relationships between the classes that an attacker can attempt to leverage. As discussed, we attribute this to a set of shared features between the whitebox and blackbox data distributions \cite{Naseer0KKP19, LeeICLR21}, as well as the propensity of DNNs to over-generalize \cite{NguyenYC15, BendaleB16}.

Importantly, we observe the correspondences are largely intuitive. 
For example, the dog, cat, and vehicle classes from distributions A and B in Test 1 show significant alignment. Similarly, in Test 2, \texttt{insect} and \texttt{beetle} both correlate with \texttt{spider}, while \texttt{train} and \texttt{boat} both align with \texttt{large-vehicle}. 
We believe this intuitive alignment to be the result of multiple factors working together. 
For one, there is undoubtedly some amount of shared ``low-frequency'' features between corresponding classes (e.g., both \texttt{beetle} and \texttt{spider} exhibit arthropodal features such as exoskeletons and thin jointed appendages). 
In addition, highly correlated classes may share similar ``high-frequency'' textures that the blackbox model is responsive to (e.g., \texttt{lizard} and \texttt{snake} have similar scaled patterns/textures on their skin). Interestingly, it has been shown that DNNs may even prioritize the learning of textures~\cite{jo2017measuring,GeirhosRMBWB19,Wang2019}, meaning the impact of any shared ``high-frequency'' features may be especially significant. A positive consequence of these intuitive alignments is that an attentive attacker may be able to successfully guess correspondences, even without probing the blackbox.

Another key observation from Figure~\ref{fig:class_correlations} is that the patterns of class correspondences are predominantly a function of the data distributions, and are relatively agnostic to the exact blackbox architecture. Observe that across the three sub-plots in either row of Figure~\ref{fig:class_correlations}, the patterns of correspondences remain consistent. From an attackers prospective, this means that once the class-relationships between the whitebox and blackbox datasets are known, they would not have to be recomputed if/when the blackbox model architecture changes. 


\vspace{-1mm}
\subsection{DINS Test 1 \& 2 attack results} \label{sec:dins_results}
\vspace{-1mm}
\textbf{DINS Test 1.}
Table~\ref{tab:test1_results} shows the transfer results in the Test 1 environment for the top eleven (target, proxy) relationships (as observed in the VGG19-Test1.B class correspondence matrix of Figure~\ref{fig:class_correlations}). For each (target, proxy) pair, we show the results of attacking with the \textit{TMIM} and \textit{FDA} algorithms across the six blackbox models. Note, we make a compromise regarding attacks that target the dog classes. From the class correspondences, both dog classes in the whitebox label space have high relation to both dog classes in the blackbox label space (not surprisingly). So, rather than selecting an individual dog to target, we configure the attack so that if the adversarial example causes the blackbox to output either \texttt{dog-terrier} or \texttt{dog-retriever} (i.e., \texttt{dog}-\textit{any}) it is considered a success. We believe this to be a reasonable trade-off for an adversary, as either way, the attack is targeting the concept of a dog.

Our first observation is that across all (target, proxy) pairs, the \textit{FDA} algorithm considerably outperforms \textit{TMIM} in both error and tSuc. 
This follows our earlier intuition that attacking in feature space may be more productive than attacking at the output layer of the whitebox in this disjoint label space setting. 
In 7 of the 11 scenarios (as shown in bold), \textit{FDA} achieves non-trivial average targeted transfer rates above $25\%$; and in 4 of these 7, the targeted transfer rates are above $50\%$. The top two (target, proxy) pairs on average are: (\texttt{dog}-\textit{any}, \texttt{dog-spaniel}) at $75.4\%/69.2\%$ error/tSuc, and (\texttt{large-vehicle}, \texttt{small-vehicle}) at $75.8\%/65.3\%$ error/tSuc. By comparison, some of the highest tSuc rates in standard ImageNet settings (where there is complete label space overlap) are about $50\%$ \cite{inkawhich_neurips20}. As such, achieving comparable transfer rates between models with disjoint label spaces is a notable result. 

Next, we remark that even though all 11 of the tested (target, proxy) relationships have strong class correspondences, not all of them reliably produce targeted adversarial examples. Specifically, 4 of the 11 \textit{FDA}-based transfers achieve $<25\%$ average tSuc, three of which are below $10\%$. This finding leads to the conclusion that high class correspondence is \textit{not} a sufficient condition for high targeted attack transferability. However, we posit that high correspondence \textit{is} necessary for reliable targeted transfer within our attack scheme (e.g., there is no reason to expect high tSuc in the (\texttt{big-game}, \texttt{bird}) scenario because the relationship between these classes is extremely weak). With that said, even though a pair may have low tSuc, the induced error rate of the attack can still be quite high. In 9 of the 11 transfers, the average error rate using \textit{FDA} is near or above $60\%$. 

\begin{table}[t]
\centering
\caption{DINS Test 2 transfers (notation = error / tSuc)}
\vspace{-2mm}
\resizebox{0.85\linewidth}{!}{
\begin{tabular}{cccc}
\toprule
    Target (Test 2.B)                & Proxy (Test 2.A)                     & TMIM-avg. &  FDA-avg.     \\ \midrule
    \textbf{\texttt{large-vehicle}} & \textbf{\texttt{train}}              & 29.5 / 8.1 &\textbf{60.8 / 44.8} \\
    \textbf{\texttt{spider}}        & \textbf{\texttt{beetle}}             &  26.4 / 5.0 & \textbf{50.7 / 27.8} \\ 
    \texttt{large-vehicle}          & \texttt{boat}               &  27.5 / 3.0 &  48.0 / 6.3 \\
    \textbf{\texttt{spider}}        & \textbf{\texttt{insect}}            &  28.2 / 7.0 &  \textbf{55.7 / 44.4} \\ 
    \texttt{fungus}                 & \texttt{fruit}              &  25.5 / 2.2 &  51.7 / 5.6 \\
    \textbf{\texttt{mustelids}}     & \textbf{\texttt{monkey}}             &  27.3 / 8.9 &  \textbf{58.9 / 24.3} \\ 
    \textbf{\texttt{house-cat}}     & \textbf{\texttt{dog-spaniel}}        &  26.0 / 5.1 &  \textbf{59.7 / 22.5} \\
    \texttt{clothing}               & \texttt{instrument} &  27.2 / 8.5 &  43.8 / 8.2 \\ 
    \texttt{fungus}                 & \texttt{crab}               &  29.4 / 2.5 &  55.7 / 8.8 \\
    \textbf{\texttt{big-game}}      & \textbf{\texttt{dog-retriever}}      &  24.3 / 1.5 &  \textbf{59.8 / 18.4} \\ \bottomrule
\end{tabular}
}
\vspace{-4mm}
\label{tab:test2_results_small}
\end{table}

\textbf{DINS Test 2.}
Table~\ref{tab:test2_results_small} shows results of attacking in the Test 2 environment for the top ten (target, proxy) relationships (as observed in the VGG19-Test2.B class-correspondence matrix of Figure~\ref{fig:class_correlations}). Here, we simply show the average error/tSuc for each (target, proxy) pair over the six blackbox models (see Appendix C for the full table). Many of the findings from Test 1 carry over to Test 2, starting with the observation that \textit{FDA} consistently outperforms \textit{TMIM} as the adversarial noise generation algorithm. In 6 of the 10 (target, proxy) pairs tested, \textit{FDA} achieves non-trivial targeted transfer rates near or above $20\%$. The top transfers on average are: (\texttt{large-vehicle}, \texttt{train}) at $60.8\%/44.8\%$ error/tSuc, and (\texttt{spider}, \texttt{insect}) at $55.7\%/44.4\%$ error/tSuc. Even in the scenarios with low targeted success, the error rates are still around $50\%$, indicating some amount of attack potency in all settings.

One intuitive takeaway when comparing results from DINS Test 1 and Test 2 is that the success rates in Test 2 are generally lower, especially among the top performing (target, proxy) pairs. We believe this is a natural consequence of the whitebox and blackbox data distributions in Test 2 having considerably less conceptual overlap (as discussed in Section 4). From our attacking perspective, such a decrease in conceptual overlap likely results in a smaller intersection between the feature sets of the whitebox and blackbox distributions for the attacks to exploit. Nevertheless, from these results we can still conclude that targeted transfer between models with no label space overlap is definitely possible, and in some cases highly effective!

\vspace{-1mm}
\subsection{Analysis: Impact of source class}
\vspace{-1mm}
The transfer results in Tables~\ref{tab:test1_results} and \ref{tab:test2_results_small} focus on the consequences of the (target, proxy) choice, and do not account for other properties of the data such as the class of the clean starting image. The displayed numbers are computed as averages over clean images from all blackbox categories besides the target class.
In this section, we investigate the impact of the clean image's class on attack transferability in an effort to further analyze conditions of high transfer.

\begin{figure}[t]
\centering
    \includegraphics[width=1\linewidth,trim={0 0 0 0},clip]{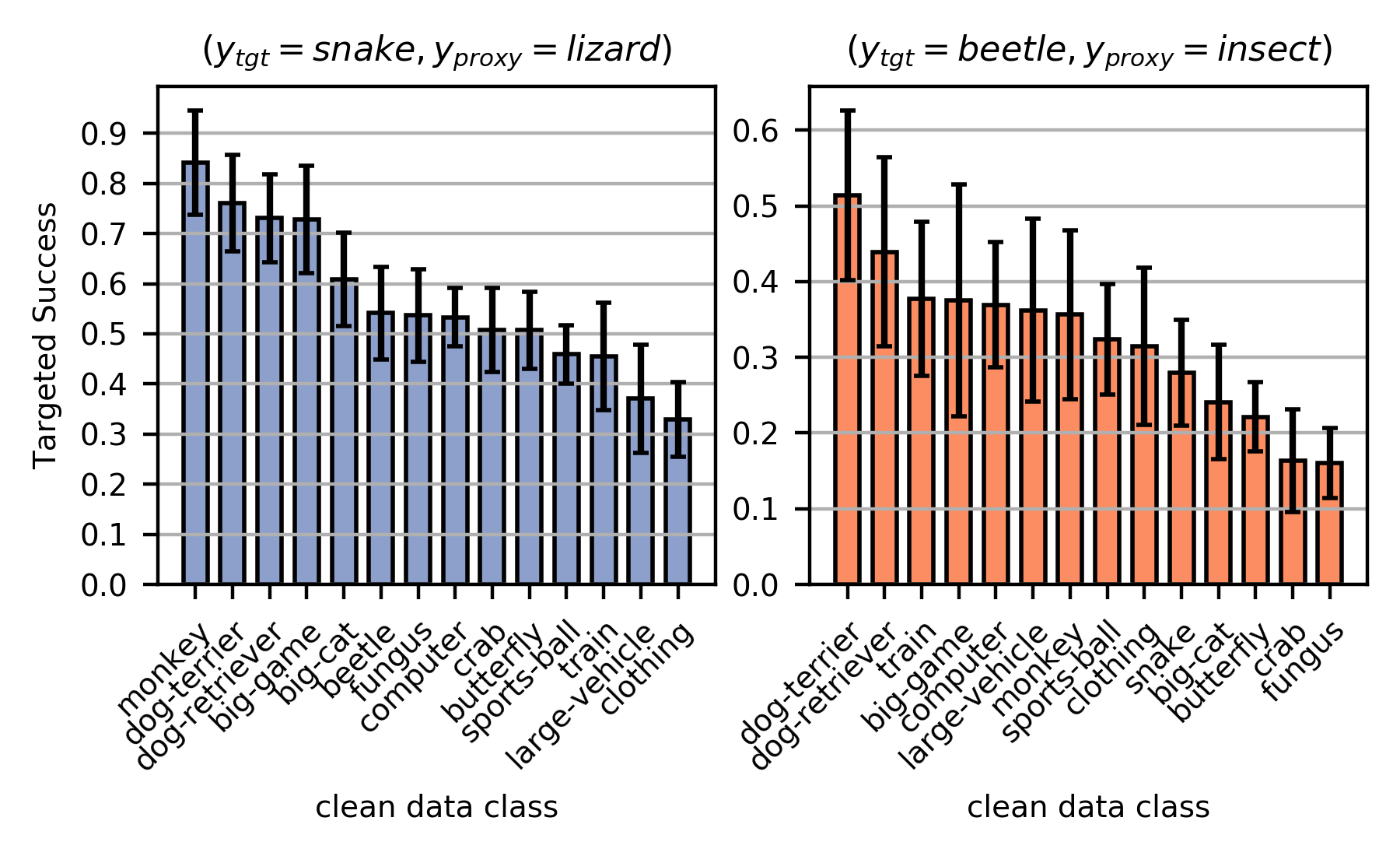}
\vspace{-8mm}
\caption{Effect of clean data class on targeted success.}
\label{fig:cleanclassimpact}
\vspace{-3mm}
\end{figure}

Figure~\ref{fig:cleanclassimpact} shows the targeted success rates of the (\texttt{snake}, \texttt{lizard}) and (\texttt{beetle}, \texttt{insect}) Test 1 transfers as a function of the clean data class (results averaged over the six Test 1 blackbox models). It is clear that attack success rates are in-fact sensitive to this information and that the patterns of class influences are not consistent across different (target, proxy) pairs (e.g., \texttt{monkey} is \textit{not} universally the best clean data class for transfers). Such behavior is likely the result of unforeseen interactions between the features of the clean data and the features of the target class. Perhaps most interestingly, notice the range of tSuc values across the classes. In the (\texttt{snake}, \texttt{lizard}) case, between \texttt{monkey} (tSuc=$84\%$) and \texttt{clothing} (tSuc=$32\%$) there is a $52\%$ range in tSuc. Similarly, in the (\texttt{beetle}, \texttt{insect}) case, the range between \texttt{dog-terrier} (tSuc=$51\%$) and \texttt{fungus} (tSuc=$16\%$) is $35\%$. Given these strong trends, this analysis also provides one method an attacker can use to adapt their attacks, to effectively maximize the chances of attack success in a given environment.

\begin{figure}[t]
\centering
 \vspace{-2mm}
    \begin{subfigure}{.9\linewidth}
    \centering
        \includegraphics[width=\linewidth,trim={0 19 270 0},clip]{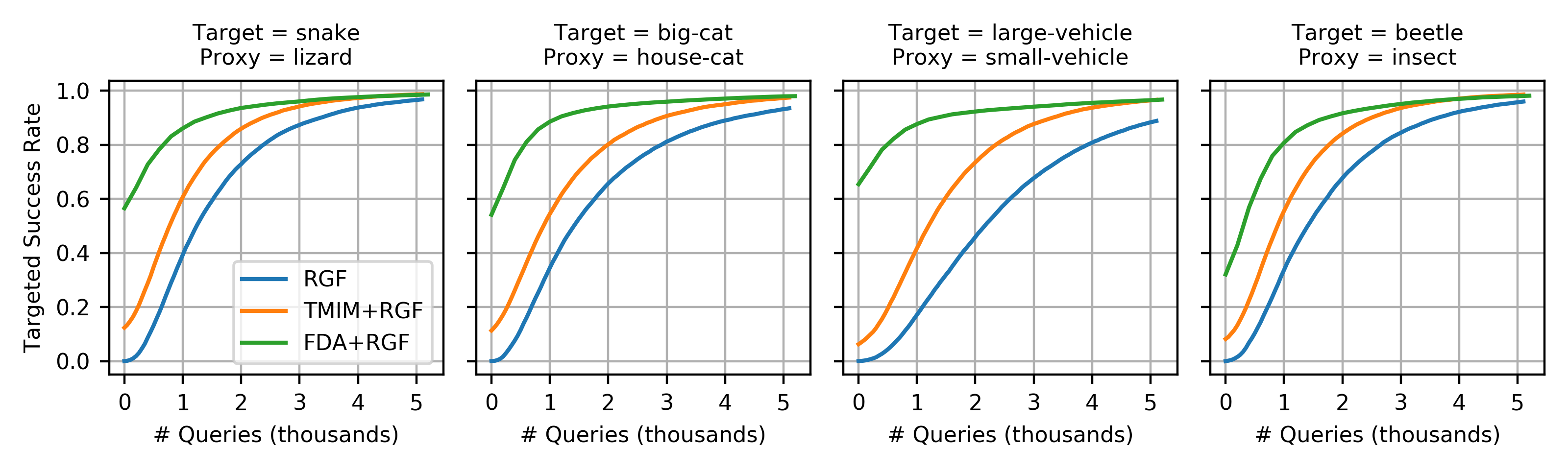}
        \vspace{-7mm}
    \end{subfigure}
    \begin{subfigure}{.9\linewidth}
    \centering
        \includegraphics[width=\linewidth,trim={0 0 270 0},clip]{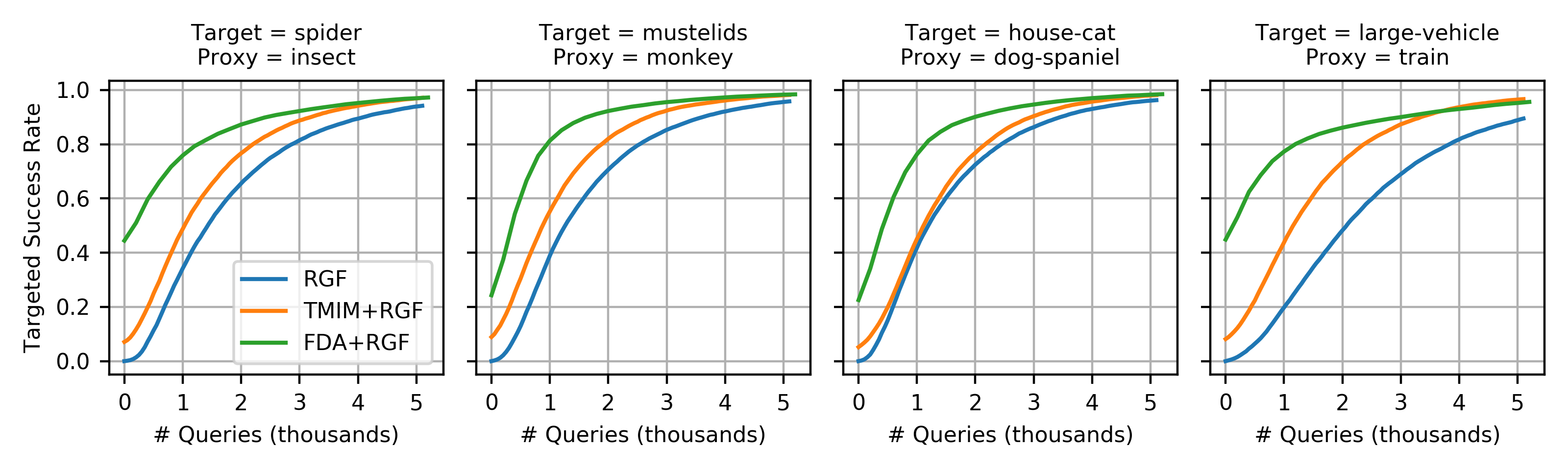}
    \end{subfigure}
     \vspace{-4mm}
\caption{Targeted attack success when integrated with the RGF query attack method.}
\vspace{-4mm}
\label{fig:query_results}
\end{figure}

\vspace{-1mm}
\subsection{Integration with query attacks} \label{sec:queries}
\vspace{-1mm}
While pure transfer attacks have the advantage of not querying the target model during attack generation, their success can be limited by the compatibility of the whitebox and blackbox models.
Although we have achieved high tSuc in many cases, it is difficult to further improve these results in a transfer-only setting.
Here, we show that by incorporating queries into our transfer attacks, we can reliably increase tSuc while requiring significantly less queries than an exclusively query-based attack \cite{ChengDPSZ19, IlyasEAL18, TuTC0ZYHC19}.
To integrate the two techniques, we first perturb for $10$ iterations using only the transfer attack. We then switch to estimating the gradient of the blackbox directly using the Random Gradient Free (\textit{RGF}) \cite{ChengDPSZ19} method and perturb until attack success (or a maximum of 5000 queries). In this way, the transfer attacks provide an initial adversarial direction that gets fine-tuned/updated with the RGF algorithm.

Figure~\ref{fig:query_results} shows the targeted success rates versus query counts ($q$) for the (\texttt{snake}, \texttt{lizard}) and (\texttt{big-cat}, \texttt{house-cat}) scenarios from DINS Test 1, and the (\texttt{spider}, \texttt{insect}) and (\texttt{mustelids}, \texttt{monkey}) scenarios from Test 2 (as averaged over the six blackbox models). The \textit{RGF} lines represent the query-only baseline, and the \textit{TMIM+RGF} and \textit{FDA+RGF} lines represent using \textit{TMIM} and \textit{FDA} to warm-start the adversarial noise, respectively. We first observe that both \textit{TMIM+RGF} and \textit{FDA+RGF} consistently outperform \textit{RGF} alone across query counts, with \textit{FDA+RGF} being the top performer. This indicates that both transfer attacks can act as powerful prior directions when used with \textit{RGF}. Interestingly, we also observe \textit{TMIM+RGF} and \textit{FDA+RGF} rapidly become more potent in the low query regime of $q\leq1000$. For example, in the (\texttt{mustelids}, \texttt{monkey}) case, the targeted success rates at $q=[0,500,1000]$ are: $[9\%, 31\%, 55\%]$ for \textit{TMIM+RGF}; $[24\%, 61\%, 81\%]$ for \textit{FDA+RGF}; and only $[0\%, 12\%, 30\%]$ for \textit{RGF}. Lastly, it is worth noting that in all four scenarios \textit{FDA+RGF} nearly reaches or exceeds $80\%$ targeted success at $q=1000$. See Appendix D for more.

\vspace{-1mm}
\subsection{ImageNet to Places365 transfers} \label{sec:inetplaces}
\vspace{-1mm}
\begin{table}[t]
\centering
\caption{ImageNet to Places365 transfers (error / tSuc)}
\vspace{-2mm}
\resizebox{\columnwidth}{!}{
\begin{tabular}{cccc}
\toprule
Target (Places365)          & Proxy (ImageNet)          & TMIM-avg.         & FDA-avg.        \\ \midrule
\texttt{83:Carousel}        & \texttt{476:Carousel}     & 63.5 / 4.9 & 93.7 / 63.0   \\
\texttt{154:Fountain}       & \texttt{562:Fountain}     & 60.2 / 8.9 & 91.0 / 71.4   \\
\texttt{40:Barn}            & \texttt{425:Barn}         & 58.1 / 2.0 & 85.1 / 20.9    \\
\texttt{300:ShoeShop}       & \texttt{788:ShoeShop}     & 58.0 / 3.5 & 94.8 / 80.8    \\
\texttt{59:Boathouse}       & \texttt{449:Boathouse}    & 55.3 / 1.0 & 86.0 / 30.6  \\
\texttt{350:Volcano}        & \texttt{980:Volcano}      & 54.1 / 0.7 & 79.5 / 32.2 \\
\texttt{72:ButcherShop}     & \texttt{467:ButcherShop}  & 59.3 / 1.5 & 90.1 / 21.3    \\
\texttt{60:Bookstore}       & \texttt{454:Bookshop}     & 56.9 / 2.4 & 90.4 / 29.6    \\ \midrule
\texttt{342:OceanDeep}      & \texttt{973:CoralReef}    & 58.2 / 3.0 & 86.2 / 37.3    \\
\texttt{76:Campsite}        & \texttt{672:MountainTent} & 56.0 / 2.0 & 88.9 / 59.5    \\
\texttt{6:AmusementArcade}  & \texttt{800:Slot}         & 62.7 / 5.1 & 92.7 / 34.9    \\
\texttt{214:Lighthouse}     & \texttt{437:Beacon}       & 54.1 / 0.5 & 76.3 / 12.7   \\
\texttt{147:FloristShop}    & \texttt{985:Daisy}        & 55.8 / 0.1 & 81.4 / 13.1    \\
\texttt{278:RailroadTrack}  & \texttt{565:FreightCar}   & 58.1 / 1.9 & 80.4 / 28.7   \\
\texttt{90:Church}          & \texttt{406:Altar}        & 62.3 / 0.6 & 93.4 / 14.1   \\
\texttt{196:JailCell}       & \texttt{743:Prison}       & 53.4 / 1.9 & 77.2 / 19.5   \\
\texttt{180:HotSpring}      & \texttt{974:Geyser}       & 53.2 / 0.1 & 83.7 / 6.6   \\
\texttt{51:Bedchamber}      & \texttt{564:FourPoster}   & 63.7 / 10.2 & 92.0 / 55.3 \\
\texttt{268:Playground}     & \texttt{843:Swing}        & 58.9 / 1.8 & 75.6 / 12.9  \\
\texttt{42:BaseballField}   & \texttt{981:Ballplayer}   & 54.2 / 0.1 & 76.6 / 22.0   \\ \bottomrule
\end{tabular}
}
\label{tab:inet_results_small}
\vspace{-4mm}
\end{table}

Our final experiment is to evaluate transfers in the more complex ImageNet to Places365 setting, as described in Section~\ref{sec:envs}. The experimental setup is nearly identical to Section~\ref{sec:setup}. For the Places365 blackbox models, we use pre-trained Wide-RN18 \cite{ZagoruykoK16}, RN50 and DN161 from the code repository of \cite{zhou2017places}. For the ImageNet whitebox models, we use an ensemble of RN50, DN121 and VGG16bn from \cite{pytorch_modelzoo}. Importantly, all of the hyperparameter tuning for the \textit{TMIM} and \textit{FDA} attacks are done using ImageNet models only, so no queries to the blackbox models are required for attack tuning. Also, all of the clean starting images we attack are correctly classified by the Places365 models, and we continue to use the $L_{\infty}~\epsilon=16/255$ noise constraint. 

First, the class correspondence matrix is constructed by forward passing the ImageNet validation set through the DN169-Places365 model and collecting the predictions. We observe many promising (target, proxy) pairs and select twenty to perform experiments with. Of the twenty, eight are classes that fall within the label space overlap (e.g., \texttt{83:Carousel} of Places365 and \texttt{476:Carousel} of ImageNet) and twelve are non-overlapping classes that have strong conceptual similarities (e.g., \texttt{342:OceanDeep} of Places365 and \texttt{973:CoralReef} of ImageNet). 
Table~\ref{tab:inet_results_small} shows the \textit{TMIM} and \textit{FDA} transfer rates for the twenty (target, proxy) pairs as averaged over the three blackbox models (see Appendix E for per-model statistics). Note, the top section are transfers between overlapping classes and the bottom section are the non-overlapping transfers. 

Consistent with previous findings, \textit{FDA} is a much more powerful noise generating algorithm than \textit{TMIM}, and is capable of inducing high error and targeted success rates for both shared and non-overlapping classes. In 9 of the 20 (target, proxy) pairs, \textit{FDA} obtains more than $30\%$ tSuc, including 5 cases that achieve over $55\%$. The top-performing (target, proxy) pair amongst the overlapping classes is (\texttt{300:ShoeShop}, \texttt{788:ShoeShop}) with $94.8\%$ error / $80.8\%$ tSuc; while the top pair amongst the non-overlapping classes is (\texttt{76:Campsite}, \texttt{672:MountainTent}) with $88.9\%$ error / $59.5\%$ tSuc. 
It is also worth noting that in 8 of the 20 transfers, \textit{FDA}-based attacks achieve over $90\%$ error, while the ``worst'' error rate amongst all twenty is still $75\%$. 
This is especially impressive, as prior work could only fool these models via query-attacks if the adversary did not have access to the Places365 dataset.
Overall, these results show that transfers between large-scale DNN models trained on distinct data distributions can be very effective. 

\vspace{-1mm}
\section{Conclusion}
\vspace{-1mm}
We describe a first-of-its-kind targeted transfer attack methodology for a new blackbox threat model where the source and target DNNs have disjoint label spaces. 
Through evaluation in difficult environments, we show that targeted attacks \textit{can} be potent even when the source and target models have highly complex data distributions with minimal conceptual overlap amongst categories.
Finally, we hope that this work inspires further research in challenging blackbox threat models that have been neglected in prior studies.


{\small
\bibliographystyle{ieee_fullname}
\bibliography{egbib}
}

\clearpage
\onecolumn
\section*{Appendix}

\subsection*{A. Super-class constructions}

Table~\ref{tab:full_class_splits} shows the construction of the 30 super-classes used in our Disjoint ImageNet Subsets (DINS) Test 1 and Test 2 experiments.
Each super-class is the composition of 5 individual ImageNet classes, based on the WordNet~\cite{miller1995wordnet} hierarchy.
We take both the training and validation data from each of the ImageNet classes, so in total this dataset has about $1300\times5\times30=195000$ training images and $50\times5\times30=7500$ validation images (we say about because not all ImageNet classes have exactly $1300$ training images).

Importantly, we do not propose that this is the only way to subset ImageNet for the purposes of constructing interesting and challenging transfer scenarios. We encourage future research to use these classes as a starting point, and to continue to build additional test environments using a similar methodological process.

\begin{table*}[h]
\centering
\caption{Construction of super-classes used in Disjoint ImageNet Subsets Tests 1 and 2}
\resizebox{\textwidth}{!}{
\begin{tabular}{ccccccc}
\toprule
Super Class        & ImageNet Components           & \multicolumn{5}{c}{ImageNet Class Names}                                                                                                  \\ \cmidrule(lr){1-1} \cmidrule(lr){2-2} \cmidrule(lr){3-7}
fish               & {[}391, 392, 393, 394, 395{]} & coho salmon               & rock beauty                    & anemone fish             & sturgeon               & garfish                  \\
bird               & {[}10, 11, 12, 13, 14{]}      & brambling                 & goldfinch                      & house finch              & junco                  & indigo bunting           \\
lizard             & {[}38, 39, 40, 41, 42{]}      & banded gecko              & iguana                         & American chameleon       & whiptail               & agama                    \\
snake              & {[}60, 61, 62, 63, 64{]}      & night snake               & boa constrictor                & rock python              & Indian cobra           & green mamba              \\
spider             & {[}72, 73, 74, 75, 76{]}      & black/gold garden spider  & barn spider                    & garden spider            & black widow            & tarantula                \\[1mm] \cdashline{1-7} \\[-2mm]
dog-hound          & {[}160, 161, 162, 163, 164{]} & Afghan hound              & basset hound                   & beagle                   & bloodhound             & bluetick                 \\
dog-terrier        & {[}179, 180, 181, 182, 183{]} & Staffordshire bullterrier & American Staffordshire terrier & Bedlington terrier       & Border terrier         & Kerry blue terrier       \\
dog-spaniel        & {[}215, 216, 217, 218, 219{]} & Brittany spaniel          & clumber spaniel                & English springer spaniel & Welsh springer spaniel & English cocker spaniel   \\
dog-retriever      & {[}205, 206, 207, 208, 209{]} & flat-coated retriever     & curly-coated retriever         & golden retriever         & Labrador retriever     & Chesapeake Bay retriever \\
house-cat          & {[}281, 282, 283, 284, 285{]} & tabby cat                 & tiger cat                      & Persian cat              & Siamese cat            & Egyptian cat             \\[1mm] \cdashline{1-7} \\[-2mm]
big-cat            & {[}286, 287, 289, 290, 292{]} & cougar                    & lynx                           & snow leopard             & jaguar                 & tiger                    \\
insect             & {[}308, 309, 310, 311, 312{]} & fly                       & bee                            & ant                      & grasshopper            & cricket                  \\
boat               & {[}510, 554, 724, 814, 871{]} & container ship            & fireboat                       & pirate ship              & speedboat              & trimaran                 \\
small-vehicle      & {[}436, 511, 609, 717, 817{]} & station wagon             & convertible                    & jeep                     & pickup                 & sports car               \\
large-vehicle      & {[}407, 779, 803, 864, 867{]} & ambulance                 & school bus                     & snowplow                 & tow truck              & trailer truck            \\[1mm] \cdashline{1-7} \\[-2mm]
turtle             & {[}33, 34, 35, 36, 37{]}      & loggerhead turtle         & leatherback turtle             & mud turtle               & terrapin               & box turtle               \\
big-game           & {[}347, 348, 349, 350, 351{]} & bison                     & ram                            & bighorn sheep            & ibex                   & hartebeest               \\
drinkware          & {[}572, 737, 898, 901, 907{]} & goblet                    & pop bottle                     & water bottle             & whiskey jug            & wine bottle              \\
train              & {[}466, 547, 565, 820, 829{]} & bullet train              & electric locomotive            & freight car              & steam locomotive       & trolley car              \\
fungus             & {[}992, 993, 994, 995, 996{]} & agaric                    & gyromitra                      & stinkhorn                & earthstar              & hen-of-the-woods         \\[1mm] \cdashline{1-7} \\[-2mm]
crab               & {[}118, 119, 120, 121, 125{]} & Dungeness crab            & rock crab                      & fiddler crab             & king crab              & hermit crab              \\
mustelids          & {[}356, 357, 359, 360, 361{]} & weasel                    & mink                           & black-footed ferret      & otter                  & skunk                    \\
instrument         & {[}402, 420, 486, 546, 594{]} & acoustic guitar           & banjo                          & cello                    & electric guitar        & harp                     \\
computer           & {[}508, 527, 590, 620, 664{]} & computer keyboard         & desktop computer               & hand-held computer       & laptop                 & monitor                  \\
fruit              & {[}948, 949, 950, 951, 954{]} & Granny Smith              & strawberry                     & orange                   & lemon                  & banana                   \\[1mm] \cdashline{1-7} \\[-2mm]
monkey             & {[}371, 372, 373, 374, 375{]} & hussar monkey             & baboon                         & macaque                  & langur                 & colobus monkey           \\
sports-ball        & {[}429, 430, 768, 805, 890{]} & baseball                  & basketball                     & rugby ball               & soccer ball            & volleyball               \\
clothing           & {[}474, 617, 834, 841, 869{]} & cardigan                  & lab coat                       & suit                     & sweatshirt             & trench coat              \\
beetle             & {[}302, 303, 304, 305, 306{]} & ground beetle             & long-horned beetle             & leaf beetle              & dung beetle            & rhinoceros beetle        \\
butterfly          & {[}322, 323, 324, 325, 326{]} & ringlet butterfly         & monarch butterfly              & cabbage butterfly        & sulphur butterfly      & lycaenid butterfly      \\ \bottomrule
\end{tabular}
}
\label{tab:full_class_splits}
\end{table*}

** The horizontal dashed lines are for visual clarity purposes only.

\newpage
\subsection*{B. Experimental setup details}

For reproducibility, we include some additional details of our experimental setup here. In this section we primarily discuss the setup as it pertains to the Disjoint ImageNet Subsets Test 1 and Test 2 environments. See Appendix E for details regarding the ImageNet to Places365 tests.

\vspace{1mm}
\noindent
\textbf{Model Training.}
The first critical step in the setup, after constructing the Test 1.A/B and Test 2.A/B dataset splits, is to train the DNN models that we will transfer attacks between. 
On Test 1.A and 2.A data we train RN50 and DN121 models to be used as whiteboxes. On Test 1.B and 2.B data we train RN34, RN152, DN169, VGG19bn, MNv2 and RXT50 models to be used as blackboxes.
All models are trained using the official PyTorch ImageNet example code from \url{https://github.com/pytorch/examples/blob/master/imagenet/main.py}.
Table~\ref{tab:model_accuracies} shows the accuracy of each model after training, as measured on the appropriate in-distribution validation split.

\begin{table}[h]
\centering
\caption{Test accuracy of models used in the DINS Test 1 and 2 experiments.}
\vspace{-2mm}
\resizebox{.25\linewidth}{!}{
\begin{tabular}{ccc}
\toprule
Train Data  & Model     & Accuracy \\ \midrule
Test-1.A    & RN50      &  94.2        \\
Test-1.A    & DN121     &  96.1        \\[1mm] \cdashline{1-3} \\[-2mm]
Test-1.B    & RN34      &  95.7        \\
Test-1.B    & RN152     &  96.2        \\
Test-1.B    & DN169     &  96.8        \\
Test-1.B    & VGG19bn   &  96.7        \\
Test-1.B    & MNv2      &  95.5        \\
Test-1.B    & RXT50     &  96.0        \\ \midrule
Test-2.A    & RN50      &  92.0        \\
Test-2.A    & DN121     &  95.0        \\[1mm] \cdashline{1-3} \\[-2mm]
Test-2.B    & RN34      &  93.6        \\
Test-2.B    & RN152     &  95.1        \\
Test-2.B    & DN169     &  95.2        \\
Test-2.B    & VGG19bn   &  95.3        \\
Test-2.B    & MNv2      &  94.0        \\
Test-2.B    & RXT50     &  93.7        \\ \bottomrule
\end{tabular}
}
\label{tab:model_accuracies}
\end{table}

\noindent
\textbf{Attack Configurations.}
For all attacks, we use a standard configuration of $L_{\infty}~\epsilon=16/255$, $\alpha=2/255$, $\mathrm{perturb\_iters}=10$, $\mathrm{momentum}=1$ when optimizing the adversarial noise \cite{ti_attack, sgm_iclr20, inkawhich_neurips20}. 
As described in \cite{inkawhich_iclr20} and \cite{inkawhich_neurips20} the Feature Distribution Attacks (FDA) require a tuning step to find a good set of attacking layers (shown as $\mathcal{L}$ in eqn.~(3)). We follow the greedy layer optimization procedure from \cite{inkawhich_neurips20} and find the following set of 4 attacking layers per whitebox model to work well:
\vspace{-1mm}
\begin{itemize}
    \itemsep-0.25em 
    \item RN50-Test1.A =  [(3,4,5), (3,4,6), (3,4,6,1), (3,4,6,2)]
    \item DN121-Test1.A = [(6,6), (6,12,10), (6,12,14), (6,12,24,12)]
    \item RN50-Test2.A =  [(3,4,5), (3,4,6), (3,4,6,1), (3,4,6,2)]
    \item DN121-Test2.A =  [(6,6), (6,12,2), (6,12,22), (6,12,24,12)]
\end{itemize}
\vspace{-1mm}
This notation comes from the way the RN50 and DN121 models are implemented in code (see \url{http://pytorch.org/vision/stable/models.html}). 
The full RN50 model has 4 layer groups with (3,4,6,3) blocks in each, and DN121 has 4 layer groups with (6,12,24,16) blocks in each. So, for example, attacking at RN50 layer (3,4,5) means we are using the feature map output from the 5$^{th}$ block in the 3$^{rd}$ layer group.
Further, we use $\eta_{RN50} = 1\mathrm{e}{-6}$, $\eta_{DN121} = 1\mathrm{e}{-5}$ to weight the contribution of the feature distance term in eqn.~(3).
As emphasized in the text, all of these hyperparameter settings are tuned by attacking between RN50-Test1.A$\leftrightarrow$DN121-Test1.A and RN50-Test2.A$\leftrightarrow$DN121-Test2.A. So, there is no dependence on querying a Test 1.B or 2.B model for hyperparameter tuning.

\vspace{1mm}
\noindent
\textbf{Attack Procedure.}
Because the images considered for attack are originally from the ImageNet validation set, each super-class has $5 \times 50 = 250$ test images. Since there are 15 super-classes in each split, and we do not consider a clean image for attack that is from the target class, we have a set of $14 \times 250 = 3,500$ images that are eligible for attack in any given (target, proxy) pair. 
As noted, we enforce that all clean starting images prior to attack are correctly classified by the blackbox models. From Table~\ref{tab:model_accuracies}, these models operate at about $\sim 95\%$ accuracy, meaning each error/tSuc number reported in Tables~\ref{tab:test1_results}, \ref{tab:test2_results_small} and \ref{tab:test2_results} is averaged over $\sim 3,300$ adversarially attacked images.

\clearpage
\subsection*{C. Full DINS Test 2 transfer results}

Table~\ref{tab:test2_results} shows the full transfer results in the Disjoint ImageNet Subsets (DINS) Test 2 environment, and is supplemental to Table~\ref{tab:test2_results_small} in the manuscript.
Note, only the numbers in the ``avg.'' column are shown in the main document (Table~\ref{tab:test2_results_small}), so the purpose of this table is to display the individual transfer rates to each blackbox model architecture.
Please refer to Section~\ref{sec:dins_results} for further discussion and analysis of these results.

\begin{table*}[h]
\centering
\caption{Transfers in the DINS Test 2 environment (notation = error / tSuc)}
\resizebox{0.9\textwidth}{!}{
\begin{tabular}{cccccccccc}
\toprule
                               &                                     &        & \multicolumn{6}{c}{Blackbox Models (Split B)}       &                              \\ \cmidrule(lr){4-9}
Target (Split B)               & Proxy (Split A)                     & Attack & RN34        & RN152       & DN169       & VGG19bn     & MNv2        & RXT50  & avg.     \\ \midrule
\multirow{2}{*}{\textbf{\texttt{large-vehicle}}} & \multirow{2}{*}{\textbf{\texttt{train}}}              & TMIM   & 29.7 / 7.0  & 23.4 / 5.3  & 27.1 / 8.3  & 27.1 / 6.4  & 33.3 / 11.7 & 36.5 / 10.0 & 29.5 / 8.1 \\
                               &                                     & FDA    & 55.4 / 35.2 & 50.7 / 39.6 & 63.9 / 52.9 & 56.5 / 41.0 & 69.9 / 57.7 & 68.1 / 42.4 & \textbf{60.8 / 44.8} \\[0.1mm] \cdashline{1-10} \\[-3mm]
\multirow{2}{*}{\textbf{\texttt{spider}}}        & \multirow{2}{*}{\textbf{\texttt{beetle}}}             & TMIM   & 26.8 / 5.3  & 20.5 / 3.0  & 23.5 / 4.2  & 25.4 / 5.2  & 30.0 / 6.9  & 32.3 / 5.3  & 26.4 / 5.0 \\
                               &                                     & FDA    & 44.9 / 23.2 & 43.6 / 27.1 & 55.6 / 21.2 & 53.6 / 36.0 & 53.5 / 31.9 & 52.9 / 27.7 & \textbf{50.7 / 27.8} \\[0.1mm] \cdashline{1-10} \\[-3mm]
\multirow{2}{*}{\texttt{large-vehicle}} & \multirow{2}{*}{\texttt{boat}}               & TMIM   & 27.6 / 3.3  & 22.2 / 2.2  & 24.7 / 1.9  & 25.7 / 2.0  & 31.1 / 5.2  & 33.6 / 3.6  & 27.5 / 3.0 \\
                               &                                     & FDA    & 48.4 / 6.9  & 40.4 / 4.1  & 49.8 / 3.0  & 41.4 / 4.2  & 51.9 / 13.8 & 55.9 / 5.9  & \textbf{48.0 / 6.3} \\[0.1mm] \cdashline{1-10} \\[-3mm]
\multirow{2}{*}{\textbf{\texttt{spider}}}        & \multirow{2}{*}{\textbf{\texttt{insect}}}             & TMIM   & 29.6 / 7.5  & 22.6 / 4.4  & 25.3 / 5.8  & 27.0 / 7.0  & 31.5 / 9.3  & 33.3 / 8.1  & 28.2 / 7.0 \\
                               &                                     & FDA    & 49.8 / 38.4 & 46.8 / 37.1 & 54.2 / 38.1 & 59.4 / 50.6 & 63.5 / 52.8 & 60.7 / 49.4 & \textbf{55.7 / 44.4} \\[0.1mm] \cdashline{1-10} \\[-3mm]
\multirow{2}{*}{\texttt{fungus}}        & \multirow{2}{*}{\texttt{fruit}}              & TMIM   & 24.8 / 1.5  & 21.6 / 2.1  & 22.7 / 2.5  & 23.7 / 1.9  & 28.1 / 2.3  & 32.1 / 2.8  & 25.5 / 2.2 \\
                               &                                     & FDA    & 49.8 / 4.2  & 55.7 / 3.4  & 59.1 / 3.1  & 43.3 / 8.1  & 49.0 / 4.7  & 53.1 / 10.2 & \textbf{51.7 / 5.6} \\[0.1mm] \cdashline{1-10} \\[-3mm]
\multirow{2}{*}{\textbf{\texttt{mustelids}}}    & \multirow{2}{*}{\textbf{\texttt{monkey}}}             & TMIM   & 27.6 / 8.9  & 22.5 / 5.0  & 27.2 / 11.8 & 25.7 / 7.0  & 30.7 / 10.7 & 30.0 / 9.7  & 27.3 / 8.9 \\
                               &                                     & FDA    & 58.3 / 28.4 & 52.5 / 21.6 & 63.3 / 29.5 & 60.5 / 17.0 & 59.6 / 22.3 & 59.4 / 27.0 & \textbf{58.9 / 24.3} \\[0.1mm] \cdashline{1-10} \\[-3mm]
\multirow{2}{*}{\textbf{\texttt{house-cat}}}     & \multirow{2}{*}{\textbf{\texttt{dog-spaniel}}}        & TMIM   & 26.2 / 4.8  & 20.1 / 3.5  & 26.1 / 5.3  & 26.3 / 8.4  & 29.6 / 5.8  & 27.9 / 2.9  & 26.0 / 5.1 \\
                               &                                     & FDA    & 58.1 / 15.0 & 52.2 / 26.4 & 63.1 / 21.8 & 60.2 / 25.8 & 65.2 / 32.1 & 59.1 / 13.8 & \textbf{59.7 / 22.5} \\[0.1mm] \cdashline{1-10} \\[-3mm]
\multirow{2}{*}{\texttt{clothing}}      & \multirow{2}{*}{\texttt{instrument}} & TMIM   & 25.0 / 8.7  & 21.1 / 7.9  & 25.6 / 9.3  & 26.3 / 8.5  & 29.9 / 8.7  & 35.0 / 7.8  & 27.2 / \textbf{8.5} \\
                               &                                     & FDA    & 41.4 / 9.9  & 24.9 / 4.7  & 39.4 / 7.5  & 43.1 / 12.9 & 55.8 / 9.3  & 58.0 / 4.9  & \textbf{43.8} / 8.2 \\[0.1mm] \cdashline{1-10} \\[-3mm]
\multirow{2}{*}{\texttt{fungus}}        & \multirow{2}{*}{\texttt{crab}}               & TMIM   & 28.8 / 1.9  & 25.8 / 1.9  & 26.6 / 3.1  & 28.9 / 3.3  & 32.7 / 2.3  & 33.6 / 2.7  & 29.4 / 2.5 \\
                               &                                     & FDA    & 52.1 / 6.6  & 53.5 / 4.2  & 58.8 / 13.6 & 54.7 / 16.4 & 60.1 / 4.2  & 55.2 / 7.9  & \textbf{55.7 / 8.8} \\[0.1mm] \cdashline{1-10} \\[-3mm]
\multirow{2}{*}{\textbf{\texttt{big-game}}}      & \multirow{2}{*}{\textbf{\texttt{dog-retriever}}}      & TMIM   & 25.0 / 1.5  & 19.3 / 0.8  & 24.5 / 2.7  & 23.7 / 0.8  & 27.9 / 2.5  & 25.7 / 0.7  & 24.3 / 1.5 \\
                               &                                     & FDA    & 56.7 / 18.7 & 52.8 / 13.0 & 61.7 / 24.0 & 56.7 / 16.4 & 64.6 / 14.4 & 66.3 / 23.9 & \textbf{59.8 / 18.4} \\ \bottomrule
\end{tabular}
}
\label{tab:test2_results}
\end{table*}

\clearpage
\subsection*{D. Additional query attack results}

\begin{figure*}[h]
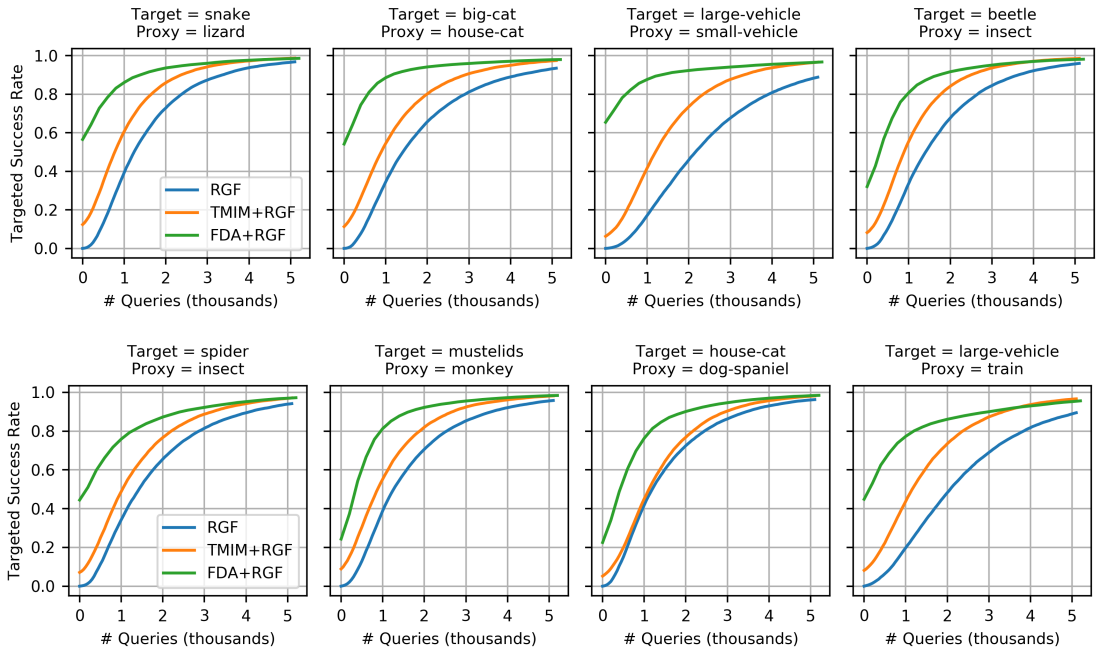

\centering
    \begin{subfigure}{.85\linewidth}
        \includegraphics[width=\linewidth,trim={0 0 0 0},clip]{figures/query_result_xv1.png}
    \end{subfigure}
    \begin{subfigure}{.85\linewidth}
        \includegraphics[width=\linewidth,trim={0 0 0 0},clip]{figures/query_result_xv2.png}
    \end{subfigure}
    \vspace{-4mm}
\caption{Targeted attack success when integrated with the RGF query attack method.}
\label{fig:query_results_full}
\end{figure*}

Figure~\ref{fig:query_results_full} is an extension of Figure~\ref{fig:query_results} in the main document, and is produced under the same experimental conditions described in Section~\ref{sec:queries}. The top row of Figure~\ref{fig:query_results_full} subplots shows the targeted success rates (tSuc) versus query counts ($q$) for the (\texttt{snake}, \texttt{lizard}), (\texttt{big-cat}, \texttt{house-cat}), (\texttt{large-vehicle}, \texttt{small-vehicle}) and (\texttt{beetle}, \texttt{insect}) scenarios from DINS Test 1.
The bottom row of subplots shows tSuc vs $q$ for the (\texttt{spider}, \texttt{insect}), (\texttt{mustelids}, \texttt{monkey}), (\texttt{house-cat}, \texttt{dog-spaniel}) and (\texttt{large-vehicle}, \texttt{train}) scenarios from DINS Test 2.
\textit{RGF} represents the query-only Random Gradient Free \cite{ChengDPSZ19} baseline attack; \textit{TMIM+RGF} represents the \textit{RGF} attack warm-started with the \textit{TMIM} transfer attack direction; and \textit{FDA+RGF} represents the \textit{RGF} attack warm-started with the \textit{FDA} transfer attack direction.
Finally, all results are averaged over the six individual blackbox models in the corresponding test environment.

\vspace{-2mm}
\vspace{-2mm}
Importantly, the results in Figure~\ref{fig:query_results_full} are consistent with those discussed in Section~\ref{sec:queries}. Both transfer attacks provide useful prior adversarial directions when integrated with the \textit{RGF} query-based attack. In practical terms, this means that using a transfer direction as a prior on the blackbox model's gradient signal, even if the source model on which the transfer direction is computed does not have the true target class in its label space, can significantly boost the query-efficiency of the RGF attack. 
To supplement Figure~\ref{fig:query_results_full}, Table~\ref{tab:query_results} shows the tSuc results of the three attacks at a query budget of $q=0/500/1000$.

\begin{table}[h]
\centering
\caption{Summary of query results at 0 / 500 / 1000 queries. This table directly supplements Figure~\ref{fig:query_results_full}.}
\vspace{-2mm}
\resizebox{0.5\columnwidth}{!}{
\begin{tabular}{cccccc}
\toprule
Test & Target & Proxy                & RGF         & TMIM+RGF     & FDA+RGF      \\ \midrule
1    & \texttt{snake} & \texttt{lizard}                & 0 / 14 / 41 & 12 / 36 / 61 & \textbf{56 / 76 / 86} \\
1    & \texttt{big-cat} & \texttt{house-cat}           & 0 / 13 / 35 & 11 / 32 / 55 & \textbf{54 / 78 / 89} \\
1    & \texttt{large-vehicle} & \texttt{small-vehicle} & 0 / 4 / 17  & 6 / 20 / 42  & \textbf{65 / 80 / 88} \\
1    & \texttt{beetle} & \texttt{insect}               & 0 / 10 / 33 & 8 / 28 / 55  & \textbf{32 / 62 / 81} \\ \midrule
2    & \texttt{spider} & \texttt{insect}               & 0 / 11 / 34 & 7 / 24 / 49  & \textbf{44 / 63 / 76} \\
2    & \texttt{large-vehicle} & \texttt{train}         & 0 / 7 / 20  & 8 / 23 / 44  & \textbf{45 / 66 / 77} \\
2    & \texttt{mustelids} & \texttt{monkey}            & 0 / 12 / 30 & 9 / 31 / 55  & \textbf{24 / 61 / 81} \\
2    & \texttt{house-cat} & \texttt{dog-spaniel}       & 0 / 15 / 43 & 5 / 21 / 45  & \textbf{22 / 55 / 76} \\ \bottomrule
\end{tabular}
}
\label{tab:query_results}
\end{table}

Notice the $q=0$ results for the \textit{TMIM+RGF} and \textit{FDA+RGF} attacks match the numbers in Tables \ref{tab:test1_results} and \ref{tab:test2_results_small}, as this is equivalent to the transfer-only (i.e., no-query) setting. Also, notice that all $q=0$ results for the RGF attack are $0\%$ tSuc. Finally, we remark that the \textit{FDA+RGF} attack is the top performer across all scenarios and can reach up to $80\%$ tSuc at $q=500$ and nearly $90\%$ tSuc at $q=1000$, depending on the particular (target, proxy) pair. Compare this to the \textit{RGF} attack alone, which can only reach about $15\%$ tSuc at $q=500$ and $43\%$ tSuc at $q=1000$ in the best case that we examine.

\clearpage
\subsection*{E. Extra materials for ImageNet to Places365 transfers}

To supplement the ImageNet to Places365 transfer results in Section~\ref{sec:inetplaces} of the main paper, here we describe some additional experimental setup details and show the expanded transfer results.

\vspace{1mm}
\noindent
\textbf{Setup.}
For all attacks, we use a standard configuration of $L_{\infty}~\epsilon=16/255$, $\alpha=2/255$, $\mathrm{perturb\_iters}=10$, $\mathrm{momentum}=1$ when optimizing the adversarial noise \cite{ti_attack, sgm_iclr20, inkawhich_neurips20}. 
Similar to the FDA layer tuning process described in Appendix B and in \cite{inkawhich_neurips20}, we tune the FDA layers for the RN50, DN121 and VGG16bn ImageNet whitebox models by attacking only amongst themselves (i.e., no queries to Places365 models required). We find the following FDA layer sets to be powerful:
\vspace{-1mm}
\begin{itemize}
    \itemsep-0.25em 
    \item RN50-ImageNet = [(3,3), (3,4), (3,4,2), (3,4,5), (3,4,6)]
    \item DN121-ImageNet = [(6,10), (6,12), (6,12,10), (6,12,18), (6,12,24,8)]
    \item VGG16bn-ImageNet = [6, 9, 11]
\end{itemize}
\vspace{-1mm}
See Appendix B for the RN50 and DN121 notation. For VGG16bn layers, the numbers indicate which convolutional layers we take the output feature maps from.
We use $\eta_{RN50} = 1\mathrm{e}{-6}$, $\eta_{DN121} = 1\mathrm{e}{-5}$, $\eta_{VGG16bn} = 1\mathrm{e}{-6}$ for the feature distance weights (see eqn.~(3)) of the RN50, DN121 and VGG16bn models, respectively.
Finally, all transfer statistics in our tests are averaged over 5000 adversarial examples, where the clean images are randomly sampled from the Places365 validation set and all are correctly classified by the Places365 blackbox models.

\vspace{1mm}
\noindent
\textbf{Expanded Results.}
Table~\ref{tab:inet_results} shows the full transfer results for the ImageNet to Places365 transfers, and is meant to supplement Table~\ref{tab:inet_results_small} in the main paper. We include this table to show the transferability results to each individual Places365 blackbox model, separately. See Section~\ref{sec:inetplaces} for a discussion on these results.


\begin{table}[h]
\centering
\caption{Attacking Places365 models via ImageNet models (notation = error / tSuc)}
\vspace{-2mm}
\resizebox{0.45\columnwidth}{!}{
\begin{tabular}{cccccc}
\toprule
                                    &                                   &        & \multicolumn{3}{c}{Blackbox Model (Places365)} \\ \cmidrule(lr){4-6}
Target (Places365)                  & Proxy (ImageNet)                  & Attack & WRN18          & RN50          & DN161         \\ \midrule
\multirow{2}{*}{\texttt{83:Carousel}}        & \multirow{2}{*}{\texttt{476:Carousel}}     & TMIM   & 65.3 / 5.4     & 64.7 / 5.4    & 60.6 / 4.0    \\
                                    &                                   & FDA    & 95.1 / 74.7    & 92.4 / 48.8   & 93.5 / 65.5   \\[1mm] \cdashline{1-6} \\[-1.75mm]
\multirow{2}{*}{\texttt{154:Fountain}}       & \multirow{2}{*}{\texttt{562:Fountain}}     & TMIM   & 63.8 / 11.6    & 59.6 / 7.6    & 57.2 / 7.6    \\
                                    &                                   & FDA    & 92.6 / 75.9    & 89.4 / 65.0   & 91.1 / 73.2   \\[1mm] \cdashline{1-6} \\[-1.75mm]
\multirow{2}{*}{\texttt{40:Barn}}            & \multirow{2}{*}{\texttt{425:Barn}}         & TMIM   & 61.4 / 2.9     & 57.6 / 1.3    & 55.3 / 1.7    \\
                                    &                                   & FDA    & 87.6 / 27.3    & 83.6 / 14.6   & 84.1 / 20.9   \\[1mm] \cdashline{1-6} \\[-1.75mm]
\multirow{2}{*}{\texttt{300:ShoeShop}}       & \multirow{2}{*}{\texttt{788:ShoeShop}}     & TMIM   & 59.5 / 2.0     & 59.4 / 6.0    & 55.0 / 2.6    \\
                                    &                                   & FDA    & 94.9 / 76.9    & 95.4 / 82.3   & 94.2 / 83.3   \\[1mm] \cdashline{1-6} \\[-1.75mm]
\multirow{2}{*}{\texttt{59:Boathouse}}       & \multirow{2}{*}{\texttt{449:Boathouse}}    & TMIM   & 58.9 / 1.8     & 55.4 / 0.4    & 51.7 / 0.8    \\
                                    &                                   & FDA    & 88.9 / 39.5    & 84.8 / 20.1   & 84.2 / 32.2   \\[1mm] \cdashline{1-6} \\[-1.75mm]
\multirow{2}{*}{\texttt{350:Volcano}}        & \multirow{2}{*}{\texttt{980:Volcano}}      & TMIM   & 56.7 / 0.8     & 54.0 / 0.5    & 51.5 / 0.8    \\
                                    &                                   & FDA    & 80.2 / 37.0    & 78.6 / 19.3   & 79.6 / 40.3   \\[1mm] \cdashline{1-6} \\[-1.75mm]
\multirow{2}{*}{\texttt{72:ButcherShop}}     & \multirow{2}{*}{\texttt{467:ButcherShop}}  & TMIM   & 62.9 / 2.1     & 58.5 / 1.2    & 56.6 / 1.3    \\
                                    &                                   & FDA    & 92.1 / 22.1    & 88.6 / 13.8   & 89.6 / 27.9   \\[1mm] \cdashline{1-6} \\[-1.75mm]
\multirow{2}{*}{\texttt{60:Bookstore}}       & \multirow{2}{*}{\texttt{454:Bookshop}}     & TMIM   & 59.4 / 2.3     & 57.5 / 2.9    & 53.9 / 1.9    \\
                                    &                                   & FDA    & 92.5 / 31.7    & 90.3 / 28.9   & 88.5 / 28.3   \\ \midrule
\multirow{2}{*}{\texttt{342:OceanDeep}}      & \multirow{2}{*}{\texttt{973:CoralReef}}    & TMIM   & 61.6 / 4.7     & 57.5 / 3.0    & 55.4 / 1.4    \\
                                    &                                   & FDA    & 87.1 / 32.3    & 84.8 / 34.9   & 86.6 / 44.8   \\[1mm] \cdashline{1-6} \\[-1.75mm]
\multirow{2}{*}{\texttt{76:Campsite}}        & \multirow{2}{*}{\texttt{672:MountainTent}} & TMIM   & 58.7 / 1.9     & 56.0 / 1.7    & 53.4 / 2.3    \\
                                    &                                   & FDA    & 90.4 / 61.4    & 86.2 / 43.6   & 90.2 / 73.5   \\[1mm] \cdashline{1-6} \\[-1.75mm]
\multirow{2}{*}{\texttt{6:AmusementArcade}}  & \multirow{2}{*}{\texttt{800:Slot}}         & TMIM   & 64.0 / 2.9     & 63.7 / 7.6    & 60.3 / 4.9    \\
                                    &                                   & FDA    & 93.1 / 30.7    & 93.4 / 43.3   & 91.6 / 30.6   \\[1mm] \cdashline{1-6} \\[-1.75mm]
\multirow{2}{*}{\texttt{214:Lighthouse}}     & \multirow{2}{*}{\texttt{437:Beacon}}       & TMIM   & 56.5 / 0.5     & 53.9 / 0.2    & 51.8 / 0.9    \\
                                    &                                   & FDA    & 79.4 / 17.0    & 74.9 / 5.6    & 74.5 / 15.4   \\[1mm] \cdashline{1-6} \\[-1.75mm]
\multirow{2}{*}{\texttt{147:FloristShop}}    & \multirow{2}{*}{\texttt{985:Daisy}}        & TMIM   & 59.0 / 0.2     & 55.5 / 0.0    & 52.8 / 0.0    \\
                                    &                                   & FDA    & 84.5 / 18.5    & 80.6 / 10.3   & 79.1 / 10.6   \\[1mm] \cdashline{1-6} \\[-1.75mm]
\multirow{2}{*}{\texttt{278:RailroadTrack}}  & \multirow{2}{*}{\texttt{565:FreightCar}}   & TMIM   & 60.6 / 2.1     & 58.4 / 1.6    & 55.3 / 2.0    \\
                                    &                                   & FDA    & 82.3 / 30.3    & 78.2 / 20.7   & 80.6 / 35.0   \\[1mm] \cdashline{1-6} \\[-1.75mm]
\multirow{2}{*}{\texttt{90:Church}}          & \multirow{2}{*}{\texttt{406:Altar}}        & TMIM   & 64.9 / 0.9     & 62.0 / 0.5    & 60.1 / 0.4    \\
                                    &                                   & FDA    & 93.9 / 20.0    & 92.7 / 7.4    & 93.6 / 14.8   \\[1mm] \cdashline{1-6} \\[-1.75mm]
\multirow{2}{*}{\texttt{196:jail\_cell}}     & \multirow{2}{*}{\texttt{743:prison}}       & TMIM   & 56.2 / 3.0     & 53.6 / 0.6    & 50.3 / 2.1    \\
                                    &                                   & FDA    & 78.4 / 25.0    & 76.3 / 7.4    & 77.0 / 26.2   \\[1mm] \cdashline{1-6} \\[-1.75mm]
\multirow{2}{*}{\texttt{180:hot\_spring}}    & \multirow{2}{*}{\texttt{974:geyser}}       & TMIM   & 56.3 / 0.1     & 53.6 / 0.1    & 49.6 / 0.2    \\
                                    &                                   & FDA    & 84.5 / 4.4     & 83.2 / 5.6    & 83.5 / 9.9    \\[1mm] \cdashline{1-6} \\[-1.75mm]
\multirow{2}{*}{\texttt{51:bedchamber}}      & \multirow{2}{*}{\texttt{564:four-poster}}  & TMIM   & 64.7 / 9.4     & 63.5 / 10.3   & 63.0 / 10.9   \\
                                    &                                   & FDA    & 92.7 / 63.3    & 91.2 / 46.1   & 92.1 / 56.6   \\[1mm] \cdashline{1-6} \\[-1.75mm]
\multirow{2}{*}{\texttt{268:playground}}     & \multirow{2}{*}{\texttt{843:swing}}        & TMIM   & 62.6 / 0.3     & 58.8 / 2.8    & 55.4 / 2.2    \\
                                    &                                   & FDA    & 77.0 / 12.1    & 75.1 / 10.3   & 74.7 / 16.3   \\[1mm] \cdashline{1-6} \\[-1.75mm]
\multirow{2}{*}{\texttt{42:baseball\_field}} & \multirow{2}{*}{\texttt{981:ballplayer}}   & TMIM   & 56.8 / 0.0     & 54.2 / 0.2    & 51.7 / 0.2    \\
                                    &                                   & FDA    & 77.7 / 15.3    & 77.0 / 23.6   & 75.1 / 27.0   \\ \bottomrule
\end{tabular}
}
\label{tab:inet_results}
\end{table}

\end{document}